\documentclass[journal]{IEEEtai}

\usepackage[colorlinks,urlcolor=blue,linkcolor=blue,citecolor=blue]{hyperref}

\usepackage{color,array}
\usepackage{graphicx}
\usepackage{amsmath}
\usepackage{amssymb}
\usepackage{bm} 
\usepackage{cite}
\usepackage{amsfonts}
\usepackage{algorithmic}

\usepackage{textcomp}
\usepackage{algorithm}
\usepackage{subfig}

\usepackage{booktabs}
\usepackage{nicematrix}
\usepackage{enumerate}
\usepackage{times}

\usepackage{multirow}

\usepackage{arydshln}
\usepackage{mathrsfs}
\usepackage{tabularx}

\begin{document}

\title{Plasticine: A Traceable Diffusion Model for Medical Image Translation}

\author{Tianyang Zhang\IEEEauthorrefmark{1}, Xinxing Cheng\IEEEauthorrefmark{1}, Jun Cheng, Shaoming Zheng, He Zhao, \\Huazhu Fu, Alejandro F Frangi, Jiang Liu, and Jinming Duan\IEEEauthorrefmark{2}

\thanks{Tianyang Zhang and Xinxing Cheng are with the Department of Computer Science, University of Birmingham, B15 2TT Birmingham, U.K. (e-mail: txz009@student.bham.ac.uk; xxc142@student.bham.ac.uk)}
\thanks{Jun Cheng is with the Institute for Infocomm Research, A*STAR, Singapore 138632. (e-mail: sam.j.cheng@gmail.com)}
\thanks{Shaoming Zheng is with Imperial College London, SW7 2AZ London, U.K. (e-mail: s.zheng22@imperial.ac.uk) }
\thanks{He Zhao is with the Department of Eye and Vision Science, University of Liverpool, L7 8TX Liverpool, U.K. (e-mail: he.zhao@liverpool.ac.uk)}
\thanks{Huazhu Fu is with the Institute of High Performance Computing, A*STAR, Singapore 138632. (e-mail: hzfu@ieee.org)}
\thanks{Alejandro F. Frangi is with the Department of Computer Science, and the Division of Informatics, Imaging and Data Sciences, School of Health Sciences, The University of Manchester, M13 9PL Manchester, U.K. (e-mail: alejandro.frangi@manchester.ac.uk)}
\thanks{Jiang Liu is with the Southern University of Science and Technology, Shen-zhen 518055, China. (e-mail: liuj@sustech.edu.cn)}
\thanks{Jinming Duan is with the University of Birmingham, B15 2TT Birmingham, U.K., and also with the University of Manchester, M13 9PL Manchester, U.K. (e-mail: jinming.duan@manchester.ac.uk).}
\thanks{\IEEEauthorrefmark{1} These authors contributed equally to this work.}
\thanks{\IEEEauthorrefmark{2} The corresponding author is Jinming Duan.}
}

\markboth{Journal of IEEE Transactions on Artificial Intelligence, Vol. 00, No. 0, Month 2020}
{First A. Author \MakeLowercase{\textit{et al.}}: Bare Demo of IEEEtai.cls for IEEE Journals of IEEE Transactions on Artificial Intelligence}

\maketitle


\begin{abstract}
  Domain gaps arising from variations in imaging devices and population distributions pose significant challenges for machine learning in medical image analysis. Existing image-to-image translation methods primarily aim to learn mappings between domains, often generating diverse synthetic data with variations in anatomical scale and shape, but they usually overlook spatial correspondence during the translation process.  For clinical applications, traceability, defined as the ability to provide pixel-level correspondences between original and translated images, is equally important. This property enhances clinical interpretability but has been largely overlooked in previous approaches. To address this gap, we propose Plasticine, which is, to the best of our knowledge, the first end-to-end image-to-image translation framework explicitly designed with traceability as a core objective. Our method combines intensity translation and spatial transformation within a denoising diffusion framework. This design enables the generation of synthetic images with interpretable intensity transitions and spatially coherent deformations, supporting pixel-wise traceability throughout the translation process.
\end{abstract}

\begin{IEEEImpStatement}
In medical imaging, existing image-to-image translation techniques often lack traceability, particularly in providing correspondences between original and translated images. This limitation compromises clinical interpretability and reduces practical applicability. To overcome these issues, we introduce Plasticine, a novel approach for image-to-image translation that prioritizes traceability. By leveraging a diffusion-based approach, Plasticine enables pixel-level synthesis while providing structural correspondences between source and target domains. The framework not only produces realistic images consistent with the target distribution but also supports the identification and interpretation of spatial and intensity changes. This significantly improves the reliability and clinical utility of medical image translation.
\end{IEEEImpStatement}


\section{Introduction}
\label{sec:intro}
\IEEEPARstart{T}{he} image-to-image translation task is to learn a mapping which  transfers the input images from one domain (source) to another (target) \cite{isola2017image,saharia2022palette},
which has a wide range of  applications in medical imaging. For example, magnetic resonance imaging (MRI) images can be translated into computed tomography (CT) to avoid CT scans, which is costly and  potentially harmful to patients \cite{yang2018unpaired}. Additionally,   the progression of some diseases, \textit{e.g.}, Alzheimer \cite{sun2020adversarial}, could also be predicted by translation between normal and abnormal cases. 
 In recent years, image translation task  has made significant progress    due to the success of the generative adversarial networks (GANs) \cite{goodfellow2014generative,denton2015deep}. GANs show remarkable capabilities in estimating distributions and sampling. For instance, MUNIT\cite{huang2018multimodal}, CycleGAN\cite{zhu2017unpaired} and StarGANs \cite{choi2018stargan,choi2020stargan} have brought promising results on the general image-to-image translation. Recently, the denoising diffusion probabilistic and score-based models\cite{sohl2015deep, ho2020denoising} have shown promising  performances on image synthesis and attracted much attention. In many tasks such as super-resolution, diffusion based methods \cite{saharia2022image} have outperformed GANs. Although there is still a lot of uncultivated space in diffusion based translation models, there are some attempts such as Plalette \cite{saharia2022palette} and DDIB\cite{su2022dual} which have both proven their efficiencies.
 

  \begin{figure}[!t]
    \centering
    \includegraphics[width=\linewidth]{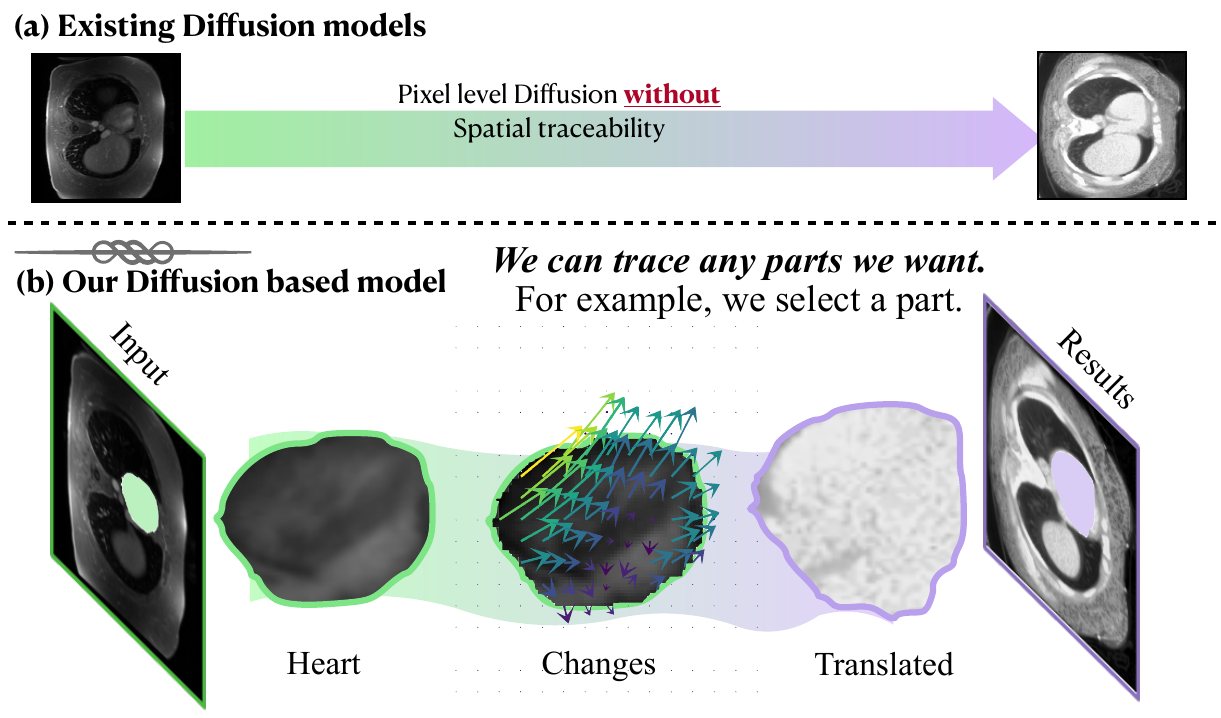}
    \caption{ Plasticine model generates high quality images with the scalability to fit the target distribution. More importantly, it has spatial traceability for clinicians to track the changes.  A demo webpage is released at \url{https://Plasticine001.github.io}.}
    \label{fig:tracebility}
    
\end{figure}

 
 
 Scalability is one of the basic requirements for image-to-image translation \cite{choi2018stargan,choi2020stargan}, as domain gaps exist not only in intensity but also in scale/shape. Many methods have been proposed to consider the scalablitiy in image translation. We find that    the traceability (finding structure correspondence and tracing changes) is another important requirement, especially for medical image translation.  Different from  applications  in natural image translation, medical image translation is used by  clinicians who often need to know the interpretability of the translation. Specifically, the correspondence of a lesion/organ
before and after the transformation would help the clinician for the disease monitoring and diagnosis. For
example in  disease progression prediction, the clinicians  want to
know how the lesions change. Therefore, tracking the changes of translation in pixel levels is crucial for medical images.
 

However, most current approaches overlook traceability. As shown in the Fig. \ref{fig:tracebility}, GAN based methods \cite{huang2018multimodal, zhu2017unpaired,choi2018stargan,choi2020stargan} achieve   image-to-image translation in   latent space without tracking spatial changes at the pixel level. Existing diffusion based methods \cite{chen2020wavegrad,saharia2022image,saharia2022palette,su2022dual}
are developed in a pixel-level translation routine, yet they fail to track the correspondence between original and synthetic images when handling the data with structure gaps.
A compromised way  to achieve this is to translate the source images without changing scales/shapes. For example,  Zhang \textit{et al.,} \cite{zhang2019noise} proposed a structure preserving GAN to translate the source image to target domain, which maintains the structures in the data for subsequent analysis. Nevertheless,   this actually led to the propagation of natural structure gaps that exists between the data from different modalities, violating the basic goal of scalability. Additionally, some works combine image-to-image translation method and registration method, such as Arar \textit{et al.} \cite{Arar_2020_CVPR}.  
However, these methods directly apply the deformations to the translation results, which face the issue that large displacement in high-frequency patterns will break the topology of the translated images and cause distortions \cite{he2022learning}.

In order to reduce the domain gaps and provide traceability (finding structure correspondence and tracing changes), we propose Plasticine, a novel  diffusion based image-to-image translation method that tackles pixel-level synthesis with both traceability and scalability. Firstly, a novel strategy that integrates the intensity translation and spatial transformation in a diffusion routine is proposed. Then we propose a  diffusion driven cross-modality spatial transformation module. It uses latent features estimated from the diffusion module to provide realistic spatial changes in the diffusion inference process.
Finally, a diffusion inference process in the proposed strategy is utilised to generate images following target distributions from source images together with the deformed structure maps. As shown in  Fig. \ref{fig:tracebility}, our model not only generates realistic images by aligning the scales/shapes to approach the target distribution, but also traces the pixel-level spatial changes. 
To evaluate the performance of the proposed method, we not only evaluate the performance of image synthesis but also  verify its traceability via segmentation applications. Additionally, we include a clinical user study with lesion/organ changes, disease progression generation, and traceability during the translation. Experimental studies on two cross-domain datasets (Retinal OCT  and Cardiac MRI)   and one cross-modality dataset (chest MRI to CT)  justify the effectiveness of the proposed methods.

In summary, we make the following \textbf{contributions}:
\begin{itemize}
    \item To the best of our knowledge, the proposed work is the first model to perform cross-domain medical image translation with the pixel-level traceability property. It benefits clinicians with the awareness of pathological progression.
    
    \item A novel Plasticine architecture that integrates intensity translation and spatial transformation is proposed in a diffusion routine. A  cross-modality spatial transformation module  driven by diffusion latent features is proposed to provide realistic spatial changes for the proposed architecture in diffusion inference.

    \item  Experimental studies on two cross-domain and one cross-modality datasets prove the superiority of Plasticine over prior arts. Specifically, we establish an adaptation segmentation application to the previous datasets to evaluate the traceability of the proposed method, which is shown in Section \ref{Sec:Setting}. Additionally, synthesis measurements and a clinical user study are also included.
\end{itemize}

\section{Background}

\begin{figure*}[!t]
    \centering \includegraphics[width=0.99\textwidth]{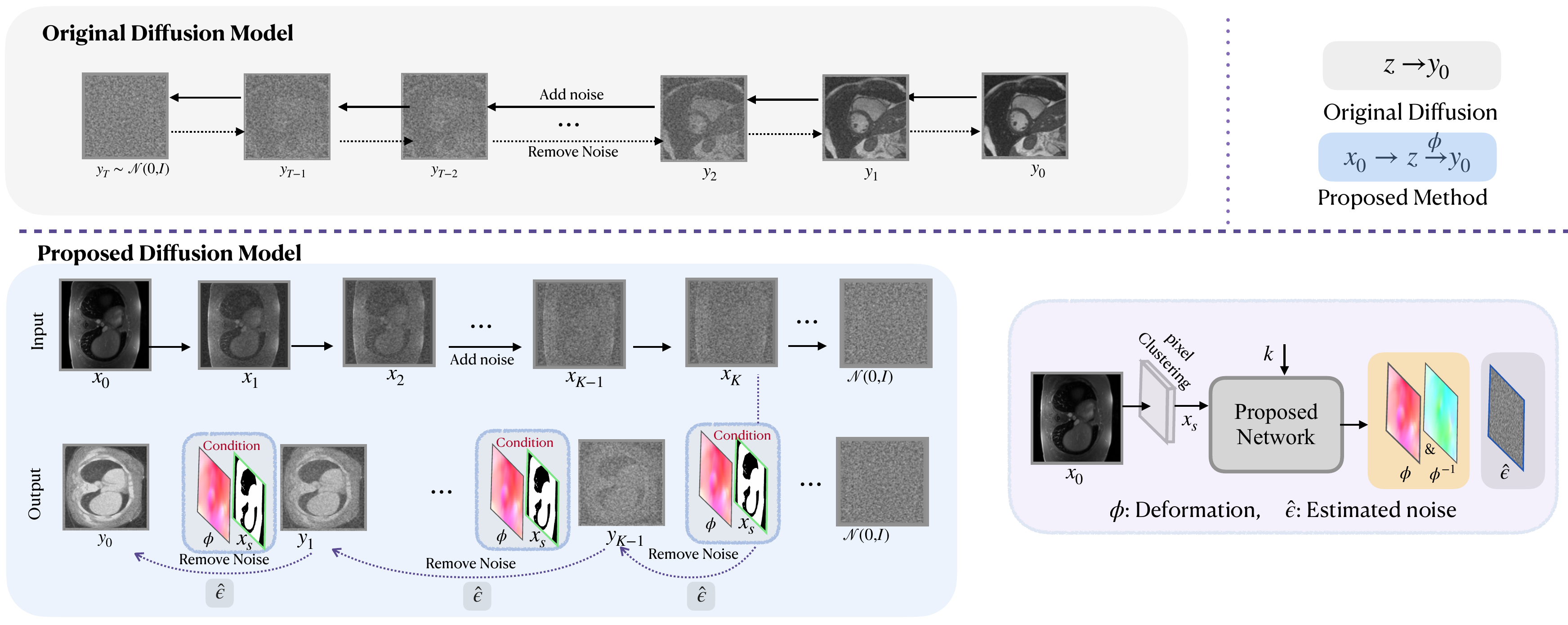}
    \caption{The figure illustrates the comparison between an original diffusion model and a proposed method named Plasticine. (a) In the original diffusion model, the image synthesis begins with random noise $\bm z$, which is progressively refined into an image $\bm y_0$. (b) In the proposed Plasticine method, the process starts with a source image $\bm x_0$. This source image is subjected to noise addition, which is then is progressively refined to to create the target image $\bm y_0$ by using deformations $\bm \phi$  basic contours $\bm x_s$, and the estimated noise $\bm{\hat{\epsilon}}$. These $\bm \phi$ and $\bm{\hat{\epsilon}}$ are obtained by a proposed network. The details of the network architecture and the complete mathematical process are provided in Methodology and Appendix Section.}
    \label{fig:top_structure}
\end{figure*}
\subsection{Image-to-Image Translation}
Image-to-image (I2I) translation obtains a mapping that   converts    an image from one domain to another. It is typically formulated as learning the conditional distribution of output given the input images. For instance, pix2pix \cite{isola2017image} learns a mapping with conditional input for translation. However, paired image data is often costly to acquire. Therefore, unpaired translation methods are more practical  and attractive for I2I translation. CycleGAN \cite{zhu2017unpaired}
and DiscoGAN \cite{kim2017learning} preserve key attributes between the input
and the translated images with a cycle consistency
loss.

\subsection{Generative Adversarial Network}
Initially proposed to provide promising fidelity and diversity in image generation, Generative Adversarial Network (GAN) \cite{goodfellow2014generative} is later used as the underlying learning paradigm of many unpaired image-to-image translation methods \cite{zhu2017unpaired,kim2017learning,huang2018multimodal,yan2019domain,choi2020stargan,zhang2024structure}. Huang \textit{et. al,}\cite{huang2018multimodal} proposes MUNIT, which separates  contents and intensity distributions in the latent space and achieves translation by switching intensity distributions. Choi \textit{et. al,}\cite{choi2020stargan} proposes StarGANv2 which translates with richer textures. For medical images,  Yan \textit{et. al,}\cite{yan2019domain} improves  CycleGAN to reduce the domain gaps   across   cardiac images from different vendors. Abu-Srhan \textit{et. al,}\cite{abu2021paired} proposes a TarGAN based on StarGAN for multi-modality image translation. Bashyam \textit{et. al,}\cite{bashyam2022deep} applies StarGANv2 on brain MRI domain translations. However, existing methods mainly consider intensity difference while fail to capture structural changes, which is vital for medical image segmentation. \textcolor{black}{Unsupervised domain adaptation (UDA) methods \cite{zhang2018translating, ren2021segmentation, chen2019synergistic} offer a promising approach to reducing domain gaps by mapping data into a different space. However, these methods typically require retraining the model or accessing latent features, which is not always feasible in clinical applications where networks are compiled into software that cannot be modified.}

\subsection{Diffusion Model}
\color{black}
In contrast to GANs\cite{goodfellow2014generative, isola2017image, zhu2017unpaired}, diffusion models \cite{sohl2015deep, ho2020denoising} belong to a family of probabilistic generative models that construct data by injecting noise progressively and denoising to reverse this process for image synthesis. 
Diffusion models provide strong inductive biases for spatial data, eliminating the need for the extensive spatial downsampling seen in other generative models operating in latent space \cite{rombach2022high}. Recently, score-based diffusion models (SBDMs) \cite{song2020score} achieved competitive or even superior image generation performance compared with GANs and thus were naturally applied to image-to-image translation. For example, SR3 \cite{saharia2022image}
and Palette \cite{saharia2022palette} learn a conditional SBDM and outperform the state-of-the-art GAN  methods on super-resolution, colorization and so on, which need paired data. 
DDIB\cite{su2022dual} transfers images by connecting source and target noising distributions through Schrödinger bridge.
EGSDE \cite{zhao2022egsde} employs an energy function pretrained on both the source
and target domains to guide the inference process of a pretrained SDE for realistic
and faithful unpaired image-to-image translation.  UNSB\cite{kim2024unpaired} reformulates the task as solving the Schrödinger bridge problem using neural stochastic differential equations, effectively unifying optimal transport and diffusion modeling.

\subsection{Spatial Transformation}

Spatial transformation models estimate the spatial mapping from a source image to a target image. Commonly, existing works encourage a diffeomorphic deformation that describes an invertible mapping such that both the forward mapping and its inverse are differentiable \cite{beg2005computing}. Conventional iterative diffeomorphic methods such as LDDMM \cite{beg2005computing}, Dartel \cite{ashburner2007fast}, ANTs \cite{avants_ANTS}, Demons\cite{vercauteren2009diffeomorphic}, and Nesterov accelerated ADMM \cite{thorley2021nesterov} are accurate but suffer from high computational costs. Building on the spatial transformer network \cite{jaderberg2015spatial}, most deep learning methods use multiple squaring and scaling as a neural layer \cite{dalca2018unsupervised, Mok_2020_CVPR, qiu2021learning,jia2025decoder} to achieve diffeomorphisms. These include VoxelMorph\cite{dalca2018unsupervised},  SYMNet\cite{Mok_2020_CVPR}, LKU-Net\cite{jia2022u}, B-spline Network \cite{qiu2021learning}, VR-Net\cite{jia2021learning}, and the most recent DiffuseMorph\cite{DiffuseMorph}, just to name a few. However, such methods are designed without considering differences in intensity distributions. \textcolor{black}{ Recent studies, \cite{Arar_2020_CVPR,chen,kong2021breaking} have applied GAN-based image translation before or after image registration to improve registration performance.  
However, these combination methods primarily aim to gain deformation rather than translation and may cause distortion problems in cases with large spatial gaps, which are discussed in \ref{ablation}.}
\section{Methodology}

\label{Sec_Method}


In this paper, we propose Plasticine, a diffusion based
image-to-image translation method that tackles pixel-level
synthesis with the spatial traceability property. 
Contrasting with the original diffusion model that generates images from normal distribution noise, the proposed diffusion translation approach synthesis image from the noised version of source image and a deformation to diminish the structural discrepancies in the diffusion inference (denoising) step. Therefore it merges the diffusion training and the reversible spatial transformation. 



An illustration of the top level structure is given in Fig. \ref{fig:top_structure}  for a better understanding of the proposed method. We define the samples from the source and target domains as $\bm x\in X$ and $\bm y \in  Y$, respectively. The basic content shapes $\bm x_s$ and $\bm y_s$ of the source and target image $\bm x$ and $\bm y$ are first obtained by an unsupervised pixel clustering method \cite{storath2015joint}. After that,  $\bm y_s$ and $\bm y_k$ ($\bm y$ embedded by the noise $\bm \epsilon$) are sent to the  diffusion noise estimation model to estimate the noise $ \bm \epsilon$, which is the normal routine of diffusion module. Then the estimated noise $\hat{\bm \epsilon}$ is utilised to generate the spatial deformation $\bm \phi$ and its reverse $\bm \phi^{-1}$ with $\bm x$ and $\bm y$ by the spatial transformation module. Additionally, the inference process applies $\bm x$ and the deformed $\bm x_s\circ \bm \phi$ to be the initial input of the diffusion reverse step. Then a iteration procedure is implemented to generate the final output $\bm {\hat{y}}$ by intermediate state $\bm {\hat{y}}_k$ and  $\bm x_s\circ \bm \phi$.

\subsection{Diffusion Intensity Translation Process}
\label{Sec_intensity}

Different from the original diffusion model, which generates images from the normal distribution noise, the proposed diffusion translation process includes a denoising process with both noised version of source image and structure alignment deformation to reduce intensity and structure gap.

\begin{figure}[!t]
    \centering
    \includegraphics[width=\linewidth]{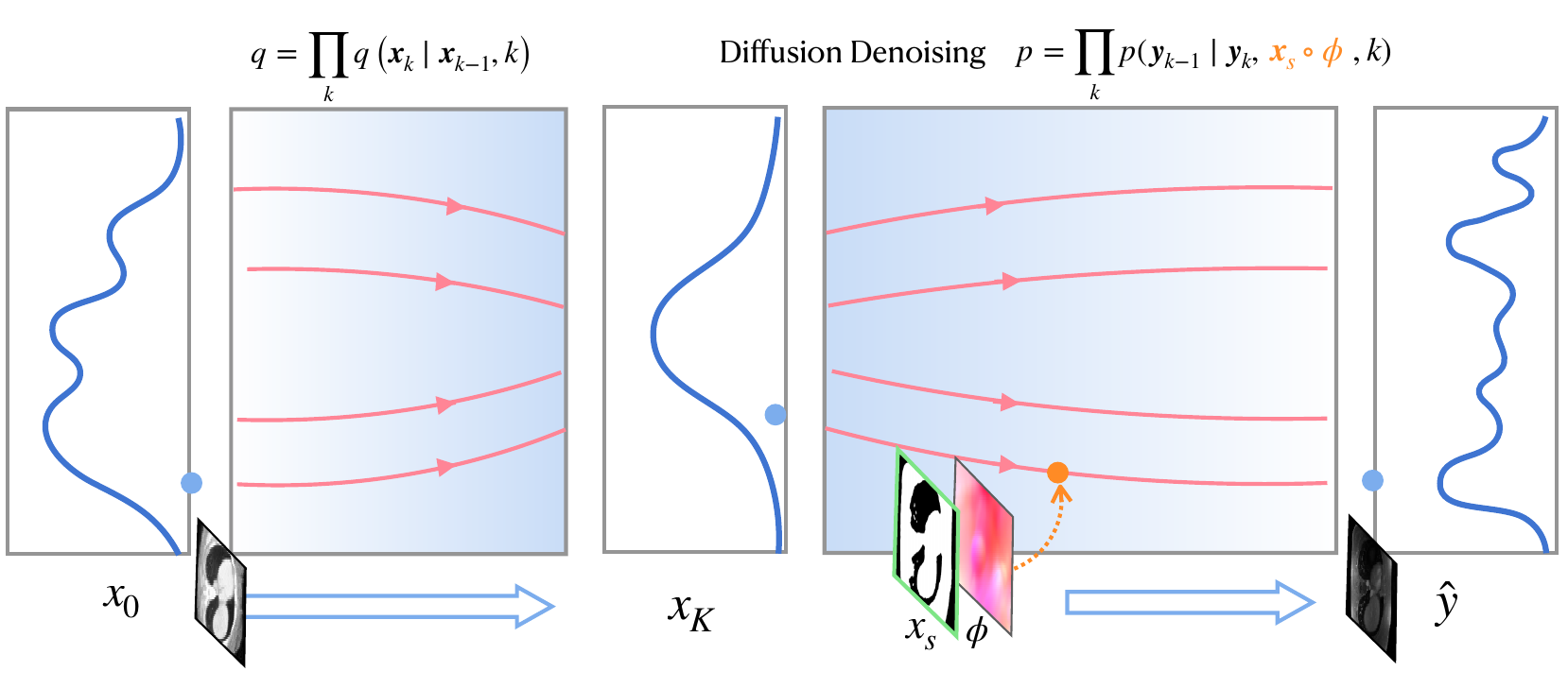}
    \caption{Illustration of the diffusion noising (add noise) and denoising (inference) process. The $\bm x_0$ and $\bm y_0$ denotes the source image $\bm x$ and the target image $\bm y$ respectively, while $\bm x_k$ and $\bm y_k$ are the noising $\bm x_0$ and $\bm y_0$ after $k$ steps. The estimated $\hat{\bm{y}}$ is obtained by the diffusion denoising process through noised input $\bm x_K$ and the deformation $\phi$.}
    \label{fig:diffusion}
\end{figure}

As shown in Fig. \ref{fig:diffusion}, the diffusion translation process includes two parts, the forward (adding noise) and the reverse (sampling synthetic image).
Firstly, the diffusion noise estimation module follows the conditional diffusion models to achieve noise estimation, where WAVEGRAD \cite{chen2020wavegrad}, diffusion based Super-Resolution\cite{saharia2022image} and Palette\cite{saharia2022palette} apply neural networks to denoise a noise-added image with the conditional input. Here, we treat the input image $\bm y_s$  as the condition, and   obtain the other input  $\bm {\tilde{ y}}$  by adding Gaussian noise to $\bm y$ with a Markovian process. Particularly, the probability of the transition  $q\left(\boldsymbol{y}_{k} \mid \boldsymbol{y}_{k-1}\right)=\mathcal{N}\left(\boldsymbol{y}_{k} ; \sqrt{\alpha_k} \boldsymbol{y}_{k-1},\left(1-\alpha_k\right) I\right)$, where $ \alpha_k$ is defined as noise schedule $\{\alpha_k\in(0,1)\}_{k=1}^K$. Mathematically, the training loss can be described as follows:
\begin{equation}
\label{eq:L_IN}
\scalebox{0.9}{
 $\mathcal{L}_{IN}=\mathbb{E}_{(\bm y_s, \bm y),\bm \epsilon,\gamma_k}\|\bm\epsilon- \bm\epsilon_{\theta_1}(\underbrace{\sqrt{\gamma_k}\bm y + \sqrt{1-\gamma_k}\bm \epsilon}_{\bm\tilde{ y}},\gamma_k,\bm y_s)\|_1$},
\end{equation}
where $\gamma_k=\prod_1^k \alpha_k$ denotes the noise level, $\bm \epsilon$ denotes the Gaussian noise $\bm \epsilon\sim \mathcal{N}(0,I)$, and $\bm \epsilon_{\theta_1}$ denotes the diffusion auto-encoder. The pseudo-code of this part is illustrated in the Appendix Algorithm \ref{alg_training_a_denoising_model}. 

For sampling synthetic image $\bm{\hat{y}}$ from the source image $\bm x_0$ and estimated noise  $\hat{\bm \epsilon} =\bm\epsilon_{\theta_1}(\bm {\hat y_k},\gamma_k,\bm x_s\circ\bm\phi)$, the image after translation $\bm{ \hat{y}} = P(\bm y_0|\bm x_0)$ could be approximated as $P(\bm y_0|\bm x_K)$. We give the proof in the Appendix equations (\ref{k_01}) and (\ref{k_02}).
Here, we embed image from the source distribution to Gaussian noise to be the initial input of the sampling process, \textit{i.e.,}, $\hat{\bm y}_K$ is estimated by $ q(\bm x_K|\bm x_0)$ which is shown as the following Markovian noising process.
\begin{equation}
\small
q(\bm x_K|\bm x_0) =\prod_{k=1}^{K}q(\bm x_k|\bm x_{k-1}) =\mathcal{N}(\bm x_K;\sqrt{\gamma_K}\bm x,(1-\gamma_K)I)
\end{equation} 
Specifically, the iterative generation process to obtain $\hat{\bm y}$ by this routine could be considered as follows:
\begin{equation}
\hat{\bm{y}}_{k-1}=\frac{1}{\sqrt{\alpha_k}}\left(\hat {\bm y_k}-\frac{1-\alpha_k}{\sqrt{1-\gamma_k}}\hat{\bm \epsilon}\right)+\sqrt{1-\alpha_k} {\bm z},
\end{equation}
where $\bm z\sim \mathcal{N}(0,I)$. 

Thus, the conditional reverse process can be formulated as
\begin{align}
    \bm{\hat y} &\sim p\left(\bm{y}_{0} \mid \bm{x}_{0}, \bm y_s\right) 
    = p\left(\bm{y}_{0} \mid \bm{x}_{K}, \bm y_s\right) 
    = p\left(\bm{y}_{0} \mid \bm{x}_{K}, \bm x_s \circ \bm \phi\right) \nonumber \\ 
    &= p\left(\bm{y}_{K-1:0} \mid \bm{x}_{K}, \bm x_s \circ \bm \phi \right) \nonumber \\ 
    &= \prod_{k=1}^{K-1} p\left(\bm{y}_k \mid \bm{y}_{k-1}, \bm x_s \circ \bm \phi, \gamma_k \right)\,
       p\left(\bm{y}_{K-1} \mid \bm{x}_{K}, \bm x_s \circ \bm \phi, \gamma_k \right).
    \label{E_all}
\end{align}

In the sampling procedure, the synthetic image $\bm{\hat{y}}$ is generated from the source image $\bm x_0$ and the estimated noise 
$\hat{\bm \epsilon} = \bm\epsilon_{\theta_1}(\hat{\bm y}_k, \gamma_k, \bm x_s \circ \bm \phi)$, 
where $\hat{\bm y}_k$ is iteratively updated as
\begin{equation}
\hat{\bm{y}}_{k-1} = \frac{1}{\sqrt{\alpha_k}}\left(\hat{\bm y}_k - \frac{1-\alpha_k}{\sqrt{1-\gamma_k}} \hat{\bm \epsilon}\right) 
+ \sqrt{1-\alpha_k}\, \bm z,
\end{equation}
with $\bm z \sim \mathcal{N}(0,I)$. 

This iterative reverse process ensures that the translated image distribution is progressively refined from Gaussian noise toward the target domain, conditioned on the source image and deformation field. Full details of the modified diffusion models and mathematical derivations are provided in the \textbf{Appendix}.

\begin{figure}[!t]
    \centering
    \includegraphics[width=\linewidth]{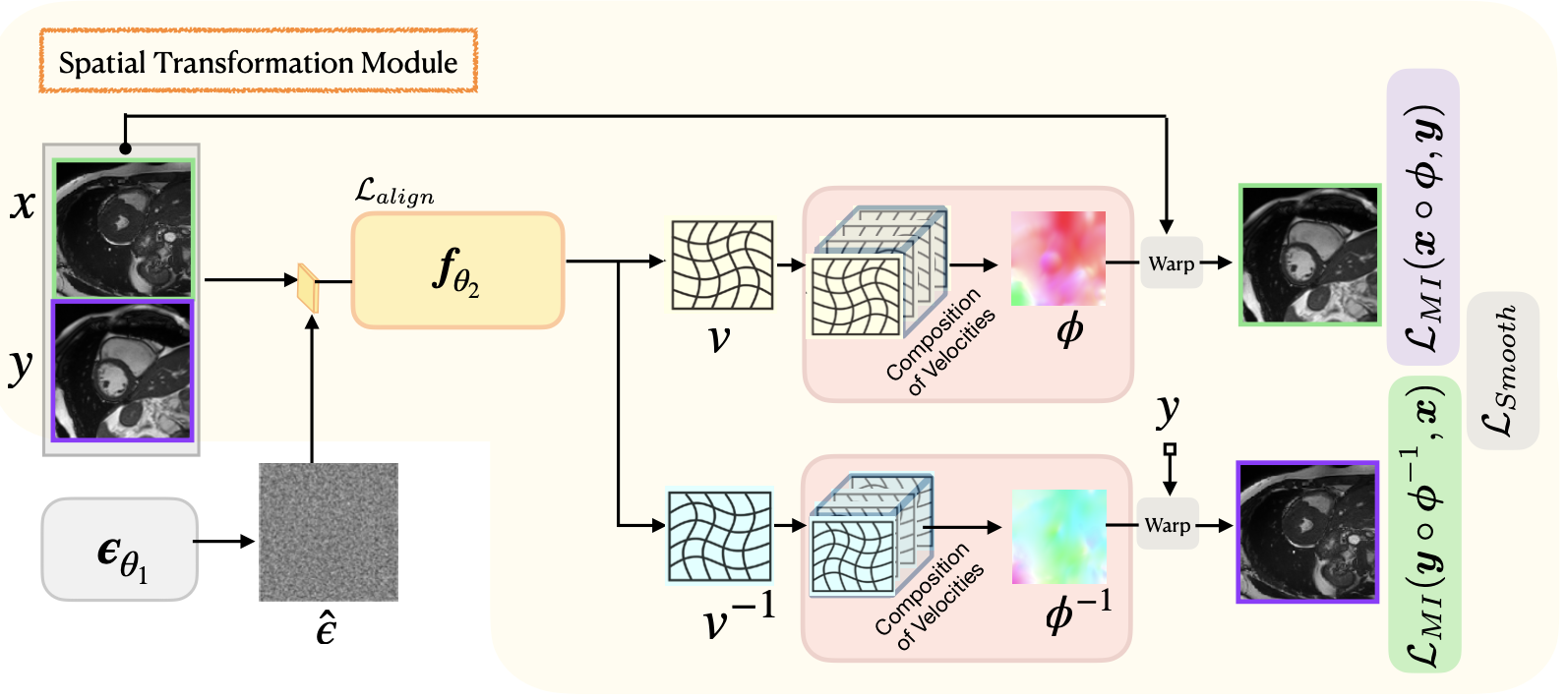}
    \caption{The training process of the deformation module. The source image $\bm x$ and target image $\bm y$ are tacked as image pairs, together with the $\bm {\hat \epsilon}$ from the diffusion auto-encoder $\bm\epsilon_{\theta_1}$, as the input of the registration module. The registration module first predicts both the initial forward and backward velocities, \textit{i.e.,} $\bm v$ and $\bm v^{-1}$, and then generates the final deformations from the velocities with a composition layer. With the computed (forward or backward) deformations, \textit{i.e.,} $\bm \phi$ and $\bm \phi^{-1}$, we can then warp an image and optimize the distances between the warped image and its corresponding reference image under a loss criterion such as Mutual Information.}
    \label{fig:spatial}
\end{figure}
\subsection{Spatial Transformation Module}
\label{spatial}
Fig.~\ref{fig:spatial} shows the spatial transformation module, and we first introduce the composition process in this module. To preserve the topology of anatomical structures and achieve realistic transformations, we require the forward and backward deformations to be diffeomorphic and mutually reversible.

We adopt a stationary ODE formulation for the deformation flow $\mathbf{\bm \phi}_t:\Omega \to \Omega$:
\[
\frac{\partial \mathbf{\bm \phi}_t}{\partial t} = \mathbf{v}(\mathbf{\bm \phi}_t), 
\quad \mathbf{\bm \phi}_0 = \mathrm{Id}, \quad t \in [0,1],
\]
where $\mathbf{v}:\Omega \to \mathbb{R}^d$ is a time-invariant velocity field and $\mathbf{\bm \phi}_0$ denotes the identity mapping. Under mild regularity assumptions ($\mathbf{v} \in C_b^1$), the Picard–Lindelöf theorem guarantees existence and uniqueness of the flow $\{\mathbf{\bm \phi}_t\}_{t\in[0,1]}$.

Furthermore, the Jacobian of the flow satisfies the variational equation
\[
\frac{\partial}{\partial t} D\mathbf{\bm \phi}_t 
= D\mathbf{v}(\mathbf{\bm \phi}_t)\, D\mathbf{\bm \phi}_t,
\]
where $D\mathbf{\bm \phi}_t \in \mathbb{R}^{d\times d}$ denotes the Jacobian of the deformation map and $D\mathbf{v}$ is the Jacobian of the velocity field. This relation describes how the local differential evolves along the flow. In particular, the determinant $J_{\mathbf{\bm \phi}_t} = \det(D\mathbf{\bm \phi}_t)$ remains positive when $D\mathbf{v}$ is bounded, thereby ensuring smoothness, invertibility, and topology preservation of the deformation.

The inverse flow $\mathbf{\bm \psi}_t = \mathbf{\bm \phi}_t^{-1}$ is generated by the negated velocity field:
\[
\frac{\partial \mathbf{\bm \psi}_t}{\partial t} = - \mathbf{v}(\mathbf{\bm \psi}_t), 
\quad \mathbf{\bm \psi}_0 = \mathrm{Id},
\]
which implies $\mathbf{\bm \phi}_1^{-1} = C(-\mathbf{v})$. Importantly, image intensities are transported consistently along the flow according to
\[
\frac{d}{dt}\big(I_0 \circ \mathbf{\bm \phi}_t^{-1}\big) = 0,
\]
ensuring brightness constancy in registration~\cite{ashburner2007fast}.

\begin{figure}[h!]
    \centering
    \includegraphics[width=\linewidth]{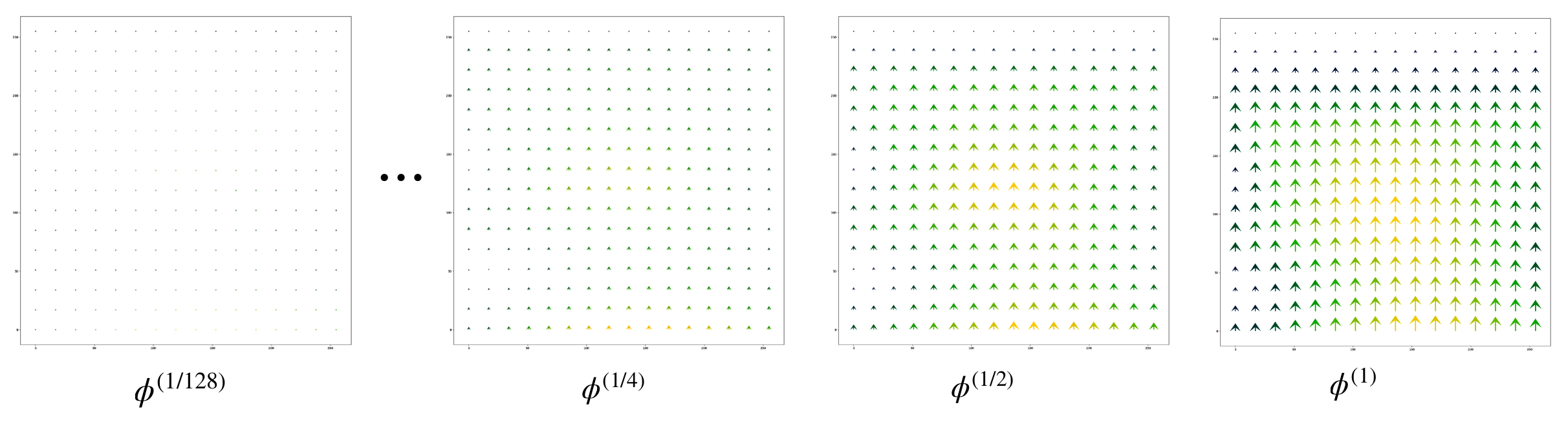}
    \caption{Illustration of the static composition process.}
    \label{fig:ill_ss}
\end{figure}

To compute the final diffeomorphism, we employ the squaring-and-scaling scheme~\cite{dalca2018unsupervised}, which approximates the exponential of the velocity field via recursive composition. We first define a small-step deformation $\mathbf{\bm \phi}^{1/2^T} \approx \mathrm{Id} + \tfrac{\mathbf{v}}{2^T}$, and then recursively square:
\[
\mathbf{\bm \phi}^{1/2^t} = \mathbf{\bm \phi}^{1/2^{t+1}} \circ \mathbf{\bm \phi}^{1/2^{t+1}}, 
\quad t = T-1,\dots,0,
\]
where $\circ$ denotes the warping (composition) operator. After $T$ steps, the forward and backward deformations are expressed as $\mathbf{\bm \phi} = C(\mathbf{v}), \;\; \mathbf{\bm \phi}^{-1} = C(-\mathbf{v})$. This construction ensures diffeomorphism and reversibility by design, numerical stability since each step is near-identity, and computational efficiency compared to dynamic velocity integration. The entire scaling-and-squaring process is illustrated in Fig.~\ref{fig:ill_ss}.

\label{sec_deformation}

We estimate the forward and backward velocities from a neural network  $\bm f_{\theta_2}$ parametrized by $\theta_2$, \textit{i.e.}, $ \bm f_{\theta_2}(\hat{\bm \epsilon}, \bm x,\bm y) =(\bm v, \bm v^{-1})$. Different from previous registration networks that only input $(\bm x,\bm y)$ \cite{dalca2018unsupervised,jia2021learning}, we include the estimated noise $\bm {\hat \epsilon}$ as an additional input cue, which contains more abstract pixel-level spatial information and is proved to be efficient in image registration \cite{DiffuseMorph}. Since no target image $\bm y$ and $\hat{\bm \epsilon}$ is available in inference, we adapt the method of feature alignment \cite{bozorgtabar2019syndemo} 
to learn the representation of target features in the latent space by training additional layers. 
Overall, to produce plausible deformations on cross-modality images, we train the network $\bm f_{\theta_2}$ with three losses. Specifically, we employ the mutual information loss \cite{qiu2021learning} $\mathcal{L}_{MI}$ as the energy term to maximize the data similarity and a smoothness loss as the regularization term $\mathcal{L}_{smooth}$ to encourage the learned velocities to be smooth. Then the feature alignment loss is adopted to predict the features map with only $\bm x$ as input. The final objective of the deformation module is: 
\begin{align} 
    \mathcal{L}_{data}&=\mathcal{L}_{MI}(\bm x\circ C(\bm v) ,\bm y)+ \mathcal{L}_{MI}(\bm x,\bm y \circ \bm C(\bm v^{-1}))\\
\mathcal{L}_{DF}&=\mathcal{L}_{data}+ \lambda_1\mathcal{L}_{smooth}+ \mathcal{L}_{align}, \label{eq:L_DF} 
\end{align}
where $\lambda_1$ is a hyper-parameter balancing the training. Further details of losses, composition operation, and warping operation are described in the \textbf{Appendix}. 

Overall, after the training phase, the pseudo-code of the inference phase can be summarised as follows.
\begin{algorithm}[h]
  \caption{Inference in $K$ iterative refinement steps}
  \begin{algorithmic}[1]
    \STATE {$\bm x_0\sim p(\bm x)$, $ \bm x_K \sim q(\bm x_K|\bm x_0).$}
    \STATE {$\hat{\bm{y}}_{K-1}=\frac{1}{\sqrt{\alpha_k}}\left(\bm{x}_K-\frac{1-\alpha_k}{\sqrt{1-\gamma_k}} \epsilon_{\theta_{1}}\left(\bm{x_s}\circ{\bm \phi}, \bm{x}_K, \gamma_k\right)\right)+\sqrt{1-\alpha_k} \bm{z}$}
    \FOR {$k = K-1, \cdots , 1$}
    \STATE {${\bm z } \sim \mathcal{N}({\bm 0},{\bm I})$ \textbf{if} $k>1$; ${\bm z =0}$ \textbf{otherwise}.}
    \STATE{$\hat{\bm{y}}_{k-1}=\frac{1}{\sqrt{\alpha_k}}\left(\hat{\bm{y}}_k-\frac{1-\alpha_k}{\sqrt{1-\gamma_k}} \epsilon_{\theta_1}\left(\bm{x_s}\circ{\bm \phi}, \hat{\bm{y}}_k, \gamma_k\right)\right)+\sqrt{1-\alpha_k} \bm{z}$}
    \ENDFOR
    \RETURN {$\hat{{\bm y}}=\hat{{\bm y}}_0$.}
  \end{algorithmic}
  \label{alg_inference_iterative_refinement_steps}
\end{algorithm}

\subsection{Implementation}

\noindent{\textbf{Network Architecture} The architecture of noise estimation network $\bm \epsilon_{\theta_1}$ is implemented using a modified version of U-Net \cite{ho2020denoising}, which is based on a conditional U-Net model for image translation \cite{saharia2022palette}. We followed this network architecture for $\bm \epsilon_{\theta_1}$, and the sample size for each input  in inference step is set to 50 to make sure we get stable synthetic images.
The architecture of the deformation module is adapted from VoxelMorph \cite{dalca2018unsupervised} that consists of a U-Net $\bm f_{\theta_2}$ to estimate velocity fields, a scaling and squaring layer to encourage diffeomorphic deformation and a spatial transformation network (STN \cite{jaderberg2015spatial}) to warp source images. We modified the final convolution layer to two output streams with the same convolution kernel $[64,2,1]$ for leaning both forward and inverse velocity fields. In addition, we customised the input alignment layer by imposing a 4-layer auxiliary convolution block with the same kernel size $[64,64,3]$ to cache the input features of $\bm y$ and $\hat{\bm \epsilon}$. The training inputs of $\bm f_{\theta_2}$ are three channels and $256\times256$ resolutions including $\bm x$, $\bm y$ and $ \hat{\bm \epsilon}$. The illustration of the two network architectures can be found in Appendix.
}\\

\noindent\textbf{Training Details} 
We applied the Adam \cite{Adam} optimizer by a learning rate of $0.0002$ and $0.0001$ with setting it to 200 and 10k linear learning rate warmup schedule for $\mathcal{L}_{DF}$ and $\mathcal{L}_{IN}$, respectively. The training time-step of the forward diffusion process applied for the diffusion intensity module was $K_{train}= 2000$ with the linearly scheduled noise level from $10^{-6}$ to $10^{-2}$. $\lambda_1$ is $0.1$, $1.0$ and $1.0$ for retinal OCT, chest MRI-CT and cardiac MRI datasets, respectively. We applied the batch size of 3 for all experiments. All training tasks were deployed on NVIDIA A100 GPU. 

\begin{figure}[!t]
     \centering
         \centering
         \includegraphics[width=\linewidth]{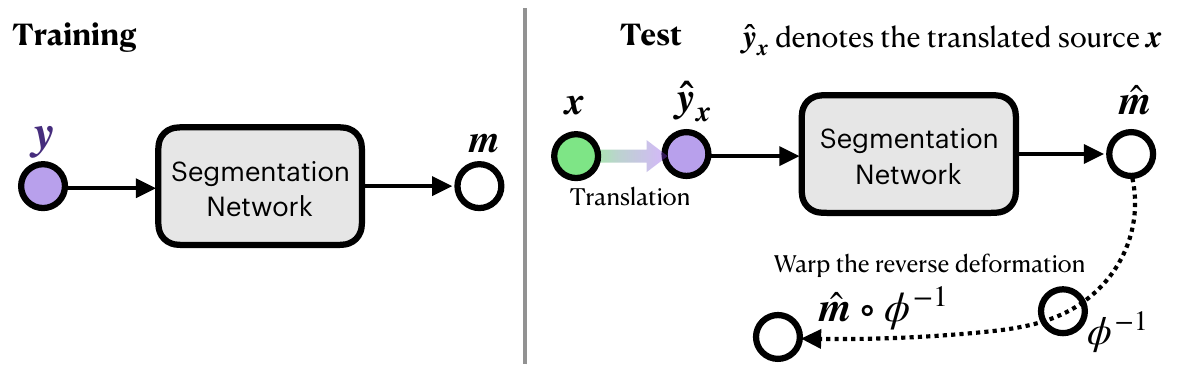}
         \label{fig:y equals x}
        \caption{Illustrations of the segmentation setting for the traceability numerical evaluation. We first train a segmentation model with the target domain images $\bm y$ and their labelled segmentation masks $\bm m$. Then, Plasticine translates the source domain images $\bm x$ into the target domain which denoted as $\hat{\bm y}_x$. The predicted masks $\hat{\bm m}$ is utilised to warp with the inverse deformation $\bm \phi^{-1}$. Next, the estimated organ shapes $\hat{\bm m} \circ \bm \phi^{-1}$ is used to evaluate the traceability of Plasticine via segmentation results.}
        \label{Fig:setting}
\end{figure}
\begin{figure}[!t]
         \centering
         \includegraphics[width=0.98\linewidth]{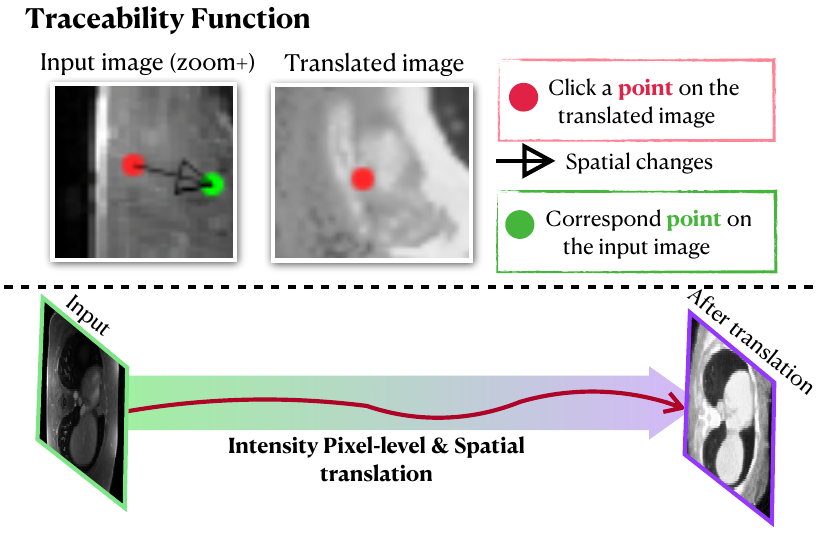}
        
        \caption{Illustrations of the translation results (MRI to CT) and the deformable changes in selected regions.  A demo Web page is released in here \url{https://Plasticine001.github.io}.  }
        \label{Fig:demo}
\end{figure}

 \begin{figure*}[t]
\center
  \includegraphics[width=\linewidth]{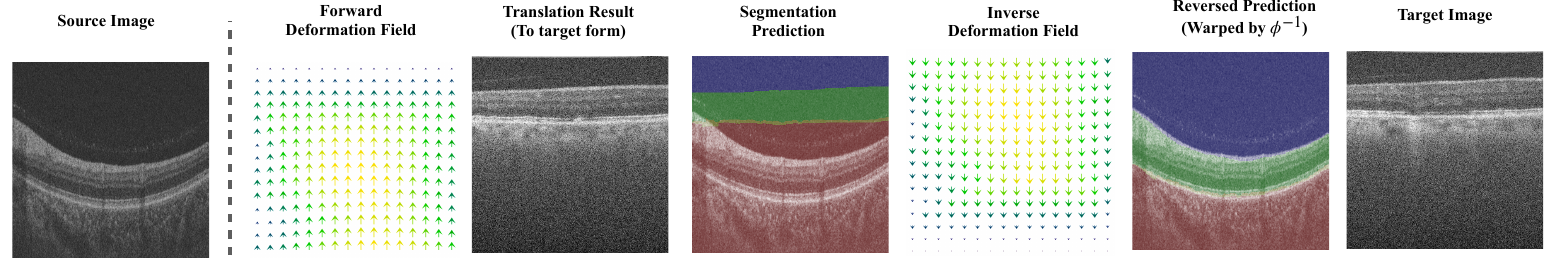}
  \caption{Illustrations of intermediate results in the segmentation setting. From left to right are source image $\bm x$, the forward deformation field $\bm \phi$,  the translated source image, the predicted result, the reverse deformation field $\bm \phi^{-1}$, the reversed predicted result and a target image.}
  \label{fig:demos_joint}
\end{figure*}

\begin{figure*}[t]
    \centering
    \includegraphics[width=\textwidth]{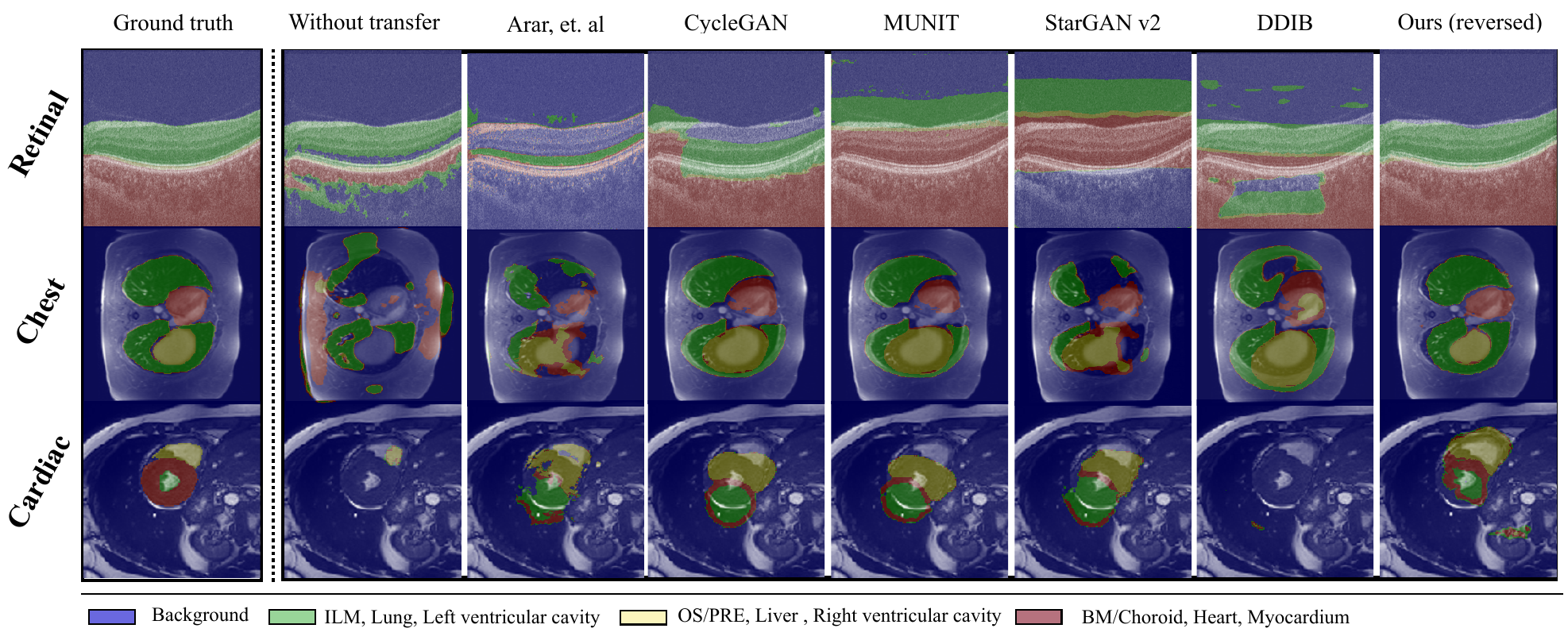}
    \caption{Visualisation of segmentation results for retinal OCT, chest MRI-CT and cardiac MRI datasets (top to bottom). Colour masks demonstrate predicted segmentation masks obtained by performing our trained segmentation model with translated images by each model. Intermediate results of Plasticine is shown in Fig. \ref{fig:demos_joint}.}
    \label{fig:seg_results}
\end{figure*}

\section{Experiments}
In this section, we experimentally study the proposed Plasticine by evaluations of  traceability and synthetic translation performance. Since it is difficult to directly evaluate the traceability, we utilise an adaptation segmentation experiment to achieve numerical evaluation, where the experimental setting is described in Section \ref{Sec:Setting}. Next, we experimentally study the image synthesis of Plasticine and state-of-the-art (SOTA)  translation models on retinal OCT, chest MRI-CT and cardiac MRI datasets. Finally, we establish an ablation study and the clinical user study. 


\subsection{Experiments Setting} 
\label{Sec:Setting}
In order to numerically analyze the traceability, we utilise an adaptation segmentation task \cite{zhang2019noise} to evaluate the proposed Plasticine and compare it to SOTA image-to-image translation methods.
Fig. \ref{Fig:setting} demonstrates the process of this adaptation segmentation. The U-Net follows the same setting described in the original papers.


To evaluate image synthesis, the image translation is performed to obtain the translated images $ \hat{\bm y}_x$ from the source domain to the target domain. Plasticine and translation-based baseline models were compared with the translated target images. Additionally, the adaptation segmentation experiments could also evaluate the synthesis performance. Because the more accurate masks it predicts, the smaller distance between the translated distribution and the target distribution is.

In clinical user studies, we expect to let clinicians track the selected regions of organs to examine their translated shape and intensity, which is shown in Fig. \ref{Fig:demo}. In order to visualize the tracking function, we built an interactive software and released a demo Web page \url{https://Plasticine001.github.io}.

\subsection{Baseline Models} 
We compare Plasticine to a number of state-of-the-art methods, which include GAN-based translation methods (such as CycleGAN\cite{zhu2017unpaired}, MUNIT\cite{huang2018multimodal}, DualGAN\cite{yi2017dualgan}, NAGAN \cite{zhang2019noise} and StarGAN-v2 \cite{choi2020stargan}), Diffusion-based translation methods (DDIB\cite{su2022dual}) and a straight forward combination method of image translation and spatial transformation proposed by  Arar, \textit{et al.} \cite{Arar_2020_CVPR} .
Particularly, all the comparison methods included in this paper follow the officially released codes and settings. Detailed descriptions of the compared methods are included in the \textbf{Appendix}. And characteristics of the comparative method are listed in the TABLE \ref{characteristic}.

 \begin{table*}[ht!]
\centering
  \caption{Segmentation results on the retinal OCT, chest MRI-CT and cardiac MRI datasets. We report the mean and highlight the best results in bold.}
\resizebox{\textwidth}{!} { 

        \begin{tabular}{lccc ccc ccc}
        \toprule[1pt]
        & \multicolumn{3}{c}{\textbf{ Retinal OCT}} & \multicolumn{3}{c}{\textbf{Chest MRI to CT}} & \multicolumn{3}{c}{\textbf{Cardiac MRI}} \\
        \cmidrule(lr){1-4} \cmidrule(lr){5-7} \cmidrule(lr){8-10}
       
    \textbf{Method}	&$Acc \uparrow$ & $Dice \uparrow$ & $mIoU \uparrow$  & $Acc \uparrow$ & $Dice \uparrow$ & $mIoU \uparrow$  & $Acc \uparrow$ & $Dice \uparrow$ & $mIoU \uparrow$  \\
    \midrule
        		Without Translation &
0.854 &
0.538  &
0.420  &

0.832  &
0.337  &
0.253  &

0.956  &
0.253 &
0.237  

\\
  Arar, \textit{et. al}\cite{Arar_2020_CVPR}   &
0.840&
0.550&
0.427&
 
0.904&
0.511&
0.388 &
 0.948
 &0.451
&0.367

 
 \\

\hdashline

        		CycleGAN \cite{zhu2017unpaired} &
0.875  &
0.624  &
0.497  &

0.930  &
0.683  &
0.549 &

\textbf{0.958}  &
0.489  &
0.399  

\\
		MUNIT \cite{huang2018multimodal} &
0.811  &
0.422  &
0.323  &

0.924  &
0.651  &
0.517  &

0.946  &
0.410  &
0.333  

\\
		DualGAN \cite{yi2017dualgan} &
0.839 &		
0.500  &
0.390  &

0.818 & 0.348 & 0.260&

0.954  &
0.275  &
0.249  

\\

		NAGAN \cite{zhang2019noise}&
0.955  &
0.748  &
0.681  &

0.841  &
0.557  &
0.432  &

0.956  &
0.271  &
0.248  

\\

		StarGAN v2 \cite{choi2020stargan} &
0.754  &		
0.312  &
0.231  &

0.907  &
0.552  &
0.421  &

0.944  &
0.439  &
0.353  
\\
 DDIB \cite{su2022dual} &
 0.881 &
 0.560 &
 0.469 &
 
 0.889 &
 0.565 &
 0.427 &
 
 0.946 &
 0.236 &
 0.223
 
 \\
  UNSB\cite{kim2024unpaired} &
 0.850 &
 0.474 &
 0.385 &
 
 0.940  &
 0.696 &
 0.578 &
 
0.957&
 0.494 &
\textbf{0.413}
 
 
 \\
\midrule
        Plasticine &
        \textbf{0.978} &
        \textbf{0.801} &
        \textbf{0.742} &
        \textbf{0.963} &
        \textbf{0.789} &
        \textbf{0.683} &
        0.957 &
        \textbf{0.502} &
       0.407
\\

      \bottomrule
        \end{tabular}
        
}
 
    \label{tab:seg_res_table}
    \end{table*}

\begin{table}
\centering
\caption{The characteristics of comparative methods.}
\begin{tabular}{lccc} 
\toprule
 \textbf{Method}     & Structure unchange & Diffusion & Deformation $\phi$ \\ 
\hline
Arar, \textit{et. al}\cite{Arar_2020_CVPR} &      $\times$             &   $\times$                   &       $\checkmark$      \\
CycleGAN \cite{zhu2017unpaired}&    $\times$                &    $\times$                 &     $\times$       \\
MUNIT \cite{huang2018multimodal}&       $\times$             &    $\times$                  &   $\times$         \\
DualGAN \cite{yi2017dualgan}&     $\times$              &  $\times$                   &     $\times$       \\
NAGAN \cite{zhang2019noise}&     $\checkmark$               &     $\times$                 &  $\times$          \\
StarGAN v2 \cite{choi2020stargan}&     $\times$               &   $\times$                  &      $\times$       \\
DDIB \cite{su2022dual} &         $\times$           &     $\checkmark$                 &   $\times$         \\
\midrule
Plasticine &       $\times$             &     $\checkmark$                 &    $\checkmark$         \\
\bottomrule
\end{tabular}
\label{characteristic}
\end{table}

\begin{figure*}[h]
    \centering
    \includegraphics[width=\textwidth]{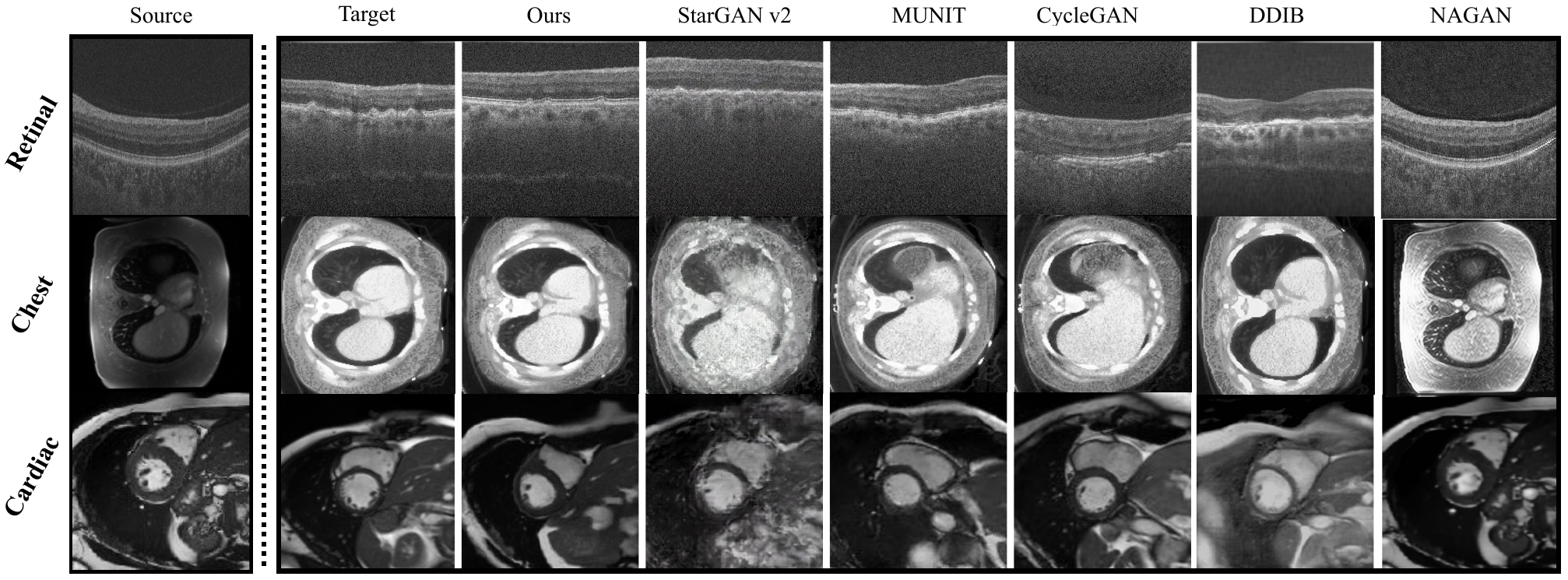}
    \caption{Visualisation of image synthesis  results for retinal OCT, chest MRI-CT and cardiac MRI datasets (top to bottom). }
    \label{fig:synthsis_results}
    \vskip -15pt
\end{figure*}

\begin{table*}[h]
\center
  \caption{Comparison of image synthesis results. The first, second and third ranks are highlighted in  \textcolor{magenta}{magenta}, \textcolor{blue}{blue} and \textcolor{cyan}{cyan} colours, respectively. }
\resizebox{\textwidth}{!} { 
        \begin{tabular}{lrrcrrcrrc}
       
        \toprule[1pt]
        & \multicolumn{3}{c}{\textbf{ Retinal OCT}} & \multicolumn{3}{c}{\textbf{Chest MRI to CT}} &
        \multicolumn{3}{c}{\textbf{Cardiac MRI }}\\
      \cmidrule(lr){1-4} \cmidrule(lr){5-7}\cmidrule(lr){8-10}
      \textbf{Method}	&$FID \downarrow$ & $SFD \downarrow$ & $D_{Bhat}$ $\downarrow$ &$FID \downarrow$ & $SFD \downarrow$ & $D_{Bhat}$ $\downarrow$ & $FID \downarrow$ & $SFD \downarrow$ & $D_{Bhat}$ $\downarrow$  \\
        \midrule




  Arar, \textit{et. al}\cite{Arar_2020_CVPR}   &
138.43&
172.99&
0.182&
 
124.53&
81.66&
0.313 &
235.49
 &128.20
&0.136

 
 \\

        CycleGAN \cite{zhu2017unpaired}  &
        \textcolor{cyan}{121.52} & 201.30 & \textcolor{cyan}{0.139} & 
        117.59 & 69.13  & \textcolor{cyan}{0.300} & 
        247.03&88.73&\textcolor{cyan}{0.109}
        
\\
		MUNIT \cite{huang2018multimodal} &
        \textcolor{blue}{119.77} & \textcolor{blue}{124.18}  &\textcolor{blue}{0.071} &
        \textcolor{cyan}{101.48} & \textcolor{magenta}{46.13} &\textcolor{magenta}{0.286} &
        \textcolor{magenta}{212.86}& \textcolor{magenta}{61.14} & \textcolor{magenta}{0.059}
\\

		DualGAN \cite{yi2017dualgan}&
        257.54 & 249.75  & 0.313 &
        256.27 & 201.43 & 0.386& 
         295.77& 170.08 & 0.259
\\
		NAGAN \cite{zhang2019noise}&
        174.12 & \textcolor{cyan}{139.53} & 0.177&
        239.70 & 187.14 & 0.320&
       245.50 & 92.06 & 0.122
\\
		StarGAN v2 \cite{choi2020stargan}&
        174.15 & 167.85 & 0.244 & 
        289.93 & 244.77 & 0.329 & 
        371.24& 249.85 & 0.185
\\
		DDIB \cite{su2022dual}&
        285.32&261.84 & 0.355&
        \textcolor{blue}{97.98}& \textcolor{cyan}{63.70}&0.339 & 
        240.45&97.23 & 0.253
\\
		UNSB\cite{kim2024unpaired} &222.34
        &207.93 &0.243 &
        266.69& 215.58& 0.337& \textcolor{blue}{217.78}
        & \textcolor{blue}{65.24}& \textcolor{blue}{0.098}
\\
\midrule
        Plasticine &
        \textcolor{magenta}{96.28} & \textcolor{magenta}{95.51} &\textcolor{magenta}{0.059}& 
        \textcolor{magenta}{ 90.00} &\textcolor{blue}{50.04} &\textcolor{blue}{0.294} & 
       \textcolor{cyan}{221.36} & \textcolor{cyan}{61.68}& 0.118\\

        \bottomrule
        \end{tabular}
}

  
    \label{tab:image_synthesis}
\end{table*}

\subsection{Datasets and Evaluation Metrics}


\noindent{\bf SINA} The SINA dataset\footnote{\url{https://people.duke.edu/~sf59/Chiu\_IOVS\_2011\_dataset.htm}} \cite{SINA} contains 220 B-scans of eyes with drusen and geographic atrophy using a spectral domain-OCT imaging system (Bioptigen). Three boundaries are manually labeled, which include the internal limiting membrane (ILM), the boundary between the outer segments and the retinal pigment epithelium (OS/RPE); and boundary and the boundary between Bruch membrane and the choroid (BM/Choroid). \medskip\\
\noindent{\bf ATLANTIS} The ATLANTIS dataset is a local dataset which contains 176 B-scans obtained from a swept-source OCT machine. Similar to SINA, the annotations of three boundaries are manually labelled. Due to the different acquisition protocols between SINA and ATLANTIS, both image structure and intensity distribution are distinct.\medskip\\
\noindent{\bf MRI-CT dataset} 
 The public multimodal chest dataset \cite{7482649,CHAOSdata2019,Clark2013} is provided by the organizers of Learn2Reg challenge\footnote{\url{https://learn2reg.grand-challenge.org}}.
 We use the scans of the chest region which contains 32 images. We select the middle slices of paired images from the volumetric MRI and CT associated with manually labelled segmentation masks (\textit{i.e.,}, heart, lung, and liver). Meanwhile, we consider the MRI and CT image scans as the source and target images, respectively.\medskip\\
\noindent{\textbf{Cardiac dataset}  Two MRI cardiac datasets the UKBiobank \cite{bernard2018deep} and the ACDC \cite{petersen2013imaging} are used in our experiments. We randomly select 100 healthy subjects from the UKBioBank and 40 patients who suffer three types of pathology (infarction, dilated cardiomyopathy, and hypertrophic cardiomyopathy) from the ACDC.
For each subject, we perform cross-domain image translation from the ACDC to the UKBioBank using the end-diastolic (ED) frames. All selected images have ground truth segmentation masks. }\medskip\\
\noindent{\bf Evaluation Metrics} To evaluate segmentation performances, we used three metrics: intersection-over-union ($mIoU$) \cite{khoreva2017simple}, the mean S{\o}rensen–Dice coefficient ($Dice$), the accuracy ($Acc$).
To evaluate translation performances, we measured the intensity distribution between synthesised and target images by using three metrics: Fr\'{e}chet inception distance ($FID$) \cite{FID}, Simplified Fréchet Distance ($SFD$) \cite{SFD} and Bhattacharyya distance $D_{Bhat}$ \cite{bhattacharyya}.

	

\subsection{Traceability Evaluation with Segmentation }

To further evaluate the translation results, we first compared the segmentation performances to measure changes in the structure for inter-subject retinal OCT, multi-modal chest MRI to CT and cross-domain cardiac MRI dataset.

\noindent
{\bf Results on segmentation.}  TABLE \ref{tab:seg_res_table} show the quantitative segmentation results of compared methods. Among the compared methods, the proposed Plasticine outperforms the previous state-of-the-art methods, i.e., GAN based methods and diffusion based method (DDIB and UNSB), and obtains the highest $Acc$, $Dice$, and $mIoU$  on retinal OCT and chest MRI-CT datasets. When compared on the cardiac MRI dataset, though the $Acc$ of Plasticine is 0.001 lower than that of CycleGAN, Plasticine outperforms all the compared methods in terms of $Dice$ with clear improvements.

In Fig. \ref{fig:seg_results}, we present the visual segmentation results of compared methods. In all three datasets, the transferred images indiscriminately generated by CycleGAN  predict the inaccurate prediction of the segmentation masks. The problems are more obvious on StarGAN-v2 and MUNIT which lose the basic organ shapes. The reason Plasticine beats CycleGAN and MUNIT is that these two models could not track structure/shape changes during translation. Similarly, diffusion-based method DDIB could not obtain a satisfactory results because it lacks spatial traceability.   The combination method proposed by Arar \textit{et al.}\cite{Arar_2020_CVPR} has texture distortion and lacks diversity which declines the translation performances, which is further discussed in Section \ref{ablation}. Particularly, the segmentation network successfully predicts organ shapes on translated images generated by Plasticine, \textit{i.e.}, heart, lung, and liver are correctly transferred to the corresponding position. This figure again suggest clear improvements brought by our Plasticine.

\subsection{Comparison on Image Synthesis}
We next compare the synthesised images generated by Plasticine with that of translation-based models. Similarly, we performed image translation on retinal OCT (ATLANTIS to SINA), chest MRI-CT (MRI to CT) and cardiac MRI (ACDC to UKBioBank).

\noindent
{\bf Results on image synthesis.} TABLE \ref{tab:image_synthesis} shows the quantitative results of image synthesis based on image distribution measurements. In the retinal OCT dataset, our Plasticine outperforms all compared models in all metrics and gains 23.49, 28.67 and 0.012 difference in $FID$, $SFD$ and $D_{Bhat}$ than that of MUNIT, respectively. For the chest MRI-CT dataset, Plasticine obtains the best $FID$ score and is close to MUNIT on $SFD$ and $D_{Bhat}$ with a difference of 3.91 and 0.008. The visual results shown in Fig. \ref{fig:synthsis_results} indicate that the images transferred by Plasticine for the chest and cardiac datasets are comparable in shape and intensity translation with that of MUNIT. Although MUNIT can generate images that are close to the target distribution, the generated images are not consistent with the original source image. Additionally, the misaligned segmentation results are not satisfying as shown in Fig. \ref{fig:seg_results}.  Specifically, the myocardium and ventricles of the heart are not accurately segmented in their proper positions by MUNIT. In the chest dataset, results obtained by MUNIT show a mismatch in the size of the labelled liver and heart.

\subsection{Ablation Study}\label{ablation}
In order to further evaluate the efficiency of the proposed model, we designed experiments to verify the necessity of some components in our proposed method.




\begin{figure}[t]
         \centering
         \includegraphics[width=\linewidth]{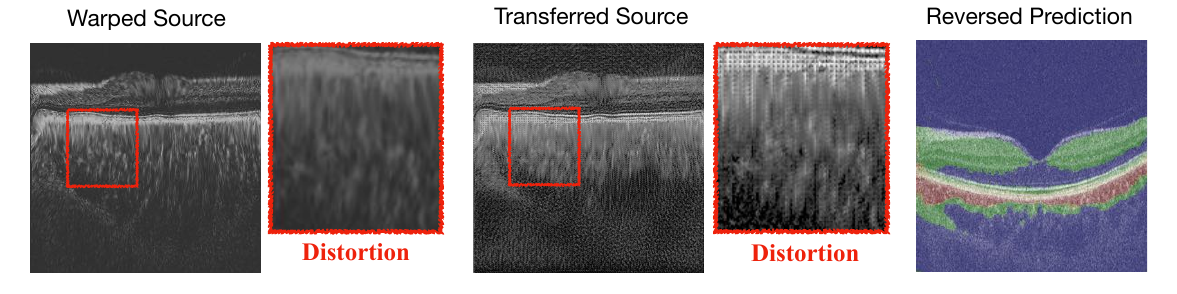}
        \caption{Illustrations of distortions in the straight forward combination method.}
        \label{Fig:composition}
\end{figure}

\begin{table}[t]
\centering
\caption{Comparison of  image synthesis results (left) and segmentation results (right). We compare Plasticine with the straightforward combination method that performs deformation followed by the translation method.}
\label{tab:ab_study_1}
\resizebox{\linewidth}{!} { 
    \begin{tabular}{lcccccc}
        \toprule[1pt]
      \textbf{Method} &$Acc \uparrow$ & $Dice \uparrow$ & $mIoU \uparrow$  &$FID \downarrow$ & $SFD \downarrow$ & $D_{Bhat}$ $\downarrow$    \\
       \midrule
        Combination & 0.796 & 0.408 & 0.305 & 289.06&248.44&0.165 \\
 

        
         Plasticine&   \textbf{0.978} &
        \textbf{0.801} &
        \textbf{0.742} & \textbf{96.28} & \textbf{95.51} &\textbf{0.059}\\
        \bottomrule[1pt]
        \end{tabular}
        }

\end{table}
\begin{figure}[t]
    \centering
    \includegraphics[width=\linewidth]{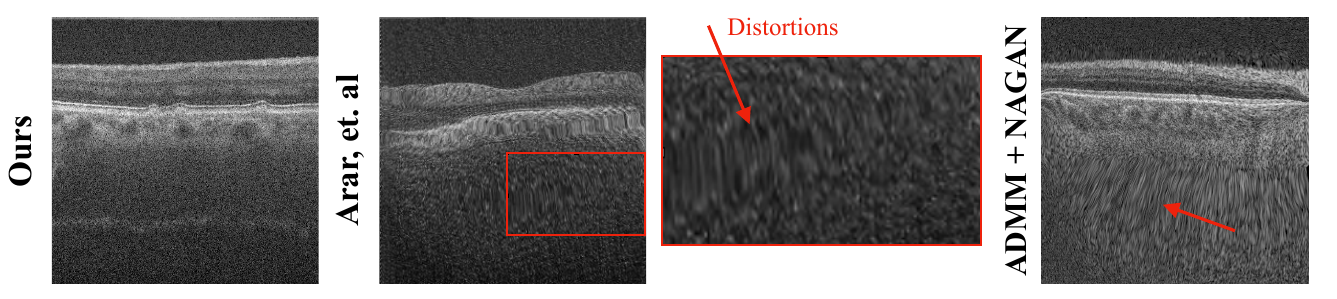}
    
    \caption{Illustration of comparison to combination methods}
    \label{fig:arar}
\end{figure}

Firstly, why should we use the proposed pipeline rather than a straight-forward combination of deformation and registration method? Here we established an experiment to evaluate this combination method (Combination), \textit{i.e.,} cross-modality deformation method followed by a structure-persevering translation method. As shown in Fig. \ref{Fig:composition} and the TABLE \ref{tab:ab_study_1}, we can see that the combination method has texture distortion which declines the translation performances, and it can not well adapt the trained segmentation network either. 
Nevertheless, our method can reach higher performances by not only generating realistic images but also capturing the spatial changes during the translation. 

Table \ref{tab:mi_comparison} reports the mutual information comparison between inputs $(x,y)$ with and without $\boldsymbol{\hat \epsilon}$. The results show that incorporating $\boldsymbol{\hat \epsilon}$ consistently improves performance, albeit slightly. Intuitively, $\boldsymbol{\hat \epsilon}$ acts as a perturbed representation of the inputs, effectively serving as an implicit data augmentation signal within the network. This encourages robustness against intensity fluctuations, acquisition noise, and small geometric distortions. Our design follows the idea in \cite{DiffuseMorph}, where noise injection has been shown to enhance spatial alignment accuracy.

\begin{table}[h!]
\centering
\caption{Mutual information (MI) comparison with and without $\hat{\boldsymbol{\epsilon}}$ on chest MR-CT dataset. Higher values indicate better alignment.}
\begin{tabular}{lcc}
\toprule
Method & MI ($x\circ \phi, y$) & MI ($y\circ \phi^{-1}, x$) \\
\midrule
with $\boldsymbol{\hat \epsilon}$   & $0.5832 \pm 0.0094$ & $0.6645 \pm 0.0057$ \\
w/o $\boldsymbol{\hat \epsilon}$    & $0.5827 \pm 0.0092$ & $0.6641 \pm 0.0057$ \\
\bottomrule
\end{tabular}
\label{tab:mi_comparison}
\end{table}

Furthermore, the study by Arar \textit{et al.}\cite{Arar_2020_CVPR} is another combination method, which has engaged in spatial transformation by using image-to-image translation method (CycleGAN) as prepossessing. Similar to the combination method we previously evaluated, the approach proposed by Arar \textit{et al.} \cite{Arar_2020_CVPR} exhibits texture distortion and lacks diversity, leading to a degradation in its translation performance. This oversight can result in distortions that compromise the fidelity of the translated images. Regrettably, subsequent attempts at rectification through registration, as indicated by He \textit{et al.} \cite{he2022learning}, do not succeed in mitigating this distortion. Our own experimental results, as outlined in TABLE \ref{tab:seg_res_table}, serve to underscore the inferior performance of the \cite{Arar_2020_CVPR} method when compared to our proposed approach. This distortion is visually evident in Fig. \ref{fig:arar}. The observed distortions could potentially arise due to significant deformations within high-frequency patterns, a consequence of the direct integration of translation and registration methodologies.

\subsection{Clinical User Study}
For further evaluating the traceability, we expect three clinicians to view and grade corresponding regions in images before and after translation. We unitised the aforementioned software in Section \ref{Sec:Setting} to organize a user study for clinicians. In particular, the clinicians need to grade the translated image, from 1 to 5, with respect to three aspects, \textit{i.e.,} the disease (\textit{e.g.,} retinal AMD Pathology Drusen) progression prediction, the fidelity of the generated images, and satisfaction with the achieved traceability. A lower score denotes better translation.
\begin{table}[t]
\centering
\caption{The average scores from three clinicians.}
\setlength\tabcolsep{1pt}
\resizebox{\linewidth}{!} { 
\begin{tabular}{lccccccccc}
        \toprule[1pt]
      \textbf{} & \multicolumn{3}{c}{\textbf{Disease Progression}} & \multicolumn{3}{c}{\textbf{Synthesis Realistic}}& \multicolumn{3}{c}{\textbf{Traceability Experience}}  \\
      \cmidrule(lr){2-4} \cmidrule(lr){5-7}\cmidrule(lr){8-10}
      \textbf{Regions} & Retinal & Chest & Cardiac& Retinal & Chest & Cardiac& Retinal & Chest & Cardiac\\
      \midrule
        Score$\downarrow$ &2.33&--&3&1.33&1.33&2.33&1.67&2&2.67\\
        \bottomrule[1pt]
         
\end{tabular}
}

\label{table:users}
\end{table}
 Table \ref{table:users} shows the results. Our method achieves plausible results, which proves the superiority of the proposed Plasticine. The detailed scores can be found in Appendix.
\section{Conclusion and Future Work}
This paper presents Plasticine, a diffusion based medical image translation method with pixel-level traceability, which integrates the spatial transformation  and the intensity translation by a denoising diffusion routine. 
Our experiments   demonstrated favorable results compared to state-of-the-art methods in various medical image translation tasks. Additionally, the adaptation segmentation experiments and a clinical user study prove  the effectiveness of the proposed method.
A limitation of the work is that it requires a structure map to be computed by traditional methods. And we would like to extend the proposed method to 3-D applications in the future.

\ifCLASSOPTIONcompsoc
  \section*{Acknowledgments}
\else
  \section*{Acknowledgment}
\fi
This research was carried out using the UK Biobank data (access number 40616). The computations described in this work were performed using the Baskerville Tier 2 HPC service. Baskerville was funded by the EPSRC and UKRI (EP/T022221/1 and EP/W032244/1) and is operated by Advanced Research Computing at the University of Birmingham. This work was supported by the UK Royal Academy of Engineering under the RAEng Chair in Emerging Technologies (INSILEX CiET1919/19), the ERC Advanced Grant UKRI Frontier Research Guarantee (INSILICO EP/Y030494/1), the UK Centre of Excellence on in-silico Regulatory Science and Innovation (UK CEiRSI) (10139527), the National Institute for Health and Care Research (NIHR) Manchester Biomedical Research Centre (BRC) (NIHR203308), the BHF Manchester Centre of Research Excellence (RE/24/130017), and the CRUK RadNet Manchester (C1994/A28701).

\appendix
\label{Sec:Appendix}
\setcounter{equation}{0}
\renewcommand{\theequation}{A.\arabic{equation}}

\section{Diffusion Models}
\subsection{Diffusion Noising (Forward)}
The forward diffusion and the reverse denoising process are two main components of diffusion models.
The forward diffusion process can be regarded as a Markovian process which repeatedly adds Gaussian noise to a data point $\bm{y}_0 \equiv \bm{y}$ over $K$ iterations:
\begin{align}
q\left(\bm{y}_{k} \mid \bm{y}_{k-1}\right) &=\mathcal{N}\left(\bm{y}_{k} ; \sqrt{\alpha_k} \bm{y}_{k-1},\left(1-\alpha_k\right) I\right) \\
q\left(\bm{y}_{1: K} \mid \bm{y}_0\right) &=\prod_{k=1}^K q\left(\bm{y}_k \mid \bm{y}_{k-1}\right)
\end{align}
where $\alpha_k$ denote hyper-parameters of the noise schedule.
The forward process with $\alpha_k$ is constructed in a manner where at $k = K$ such that ${\bm y}_K$ is virtually indistinguishable from Gaussian noise.
Note that we marginalise the forward process at each step:
\begin{equation}
q\left(\bm{y}_k \mid \bm{y}_0\right)=\mathcal{N}\left(\bm{y}_k ; \sqrt{\gamma_k} \bm{y}_0,\left(1-\gamma_k\right) I\right),
\end{equation}
where $\gamma_k=\prod_k \alpha_k$.

We can also use the Gaussian parameterisation of the forward process to form the posterior distribution of ${\bm y}_{k-1}$ for given $({\bm y}_0, {\bm y}_k)$:
\begin{equation}
q\left(\bm{y}_{k-1} \mid \bm{y}_0, \bm{y}_k\right)=\mathcal{N}\left(\bm{y}_{k-1} \mid \bm{\mu}, \sigma^2 \bm{I}\right),
\end{equation}
where $\bm{\mu}=\frac{\sqrt{\gamma_{k-1}}\left(1-\alpha_k\right)}{1-\gamma_k} \bm{y}_0+\frac{\sqrt{\alpha_k}\left(1-\gamma_{k-1}\right)}{1-\gamma_k} \bm{y}_k$ and $\sigma^2=\frac{\left(1-\gamma_{k-1}\right)\left(1-\alpha_k\right)}{1-\gamma_k}$.

\subsection{Training}

Plasticine trains a network to estimate the noise for the further reverse process. Specifically,  we adopt a neural network to estimate the noise  $\hat{\bm \epsilon} =\bm\epsilon_{\theta_1}(\bm {\tilde y},\gamma_k,\bm y_s)$ 
 from a given noise embedded target image $
\tilde{\bm y}$,
\begin{equation}
\widetilde{\bm{y}}=\sqrt{\gamma} \bm{y}_0+\sqrt{1-\gamma} \bm{\epsilon}, \bm{\epsilon} \sim \mathcal{N}(0, I),
\end{equation}
where $\bm y_s$ denotes the structure map of $\bm y_0$, , and $\gamma$ is the current noise level.
The training of noise  prediction is minimising the following loss.
\begin{equation}
 \mathcal{L}_{IN}=\mathbb{E}_{(\bm y_s, \bm y),\bm \epsilon,\gamma_k}\|\bm\epsilon- \bm\epsilon_{\theta_1}(\underbrace{\sqrt{\gamma_k}\bm y + \sqrt{1-\gamma_k}\bm \epsilon}_{\bm\tilde{ y}},\gamma_k,\bm y_s)\|_1
\end{equation}

\begin{algorithm}[h]
  \caption{Training the noise estimation model $ \bm\epsilon_{\theta_1}$}
  \begin{algorithmic}[1]
    \REPEAT
    \STATE {${\bm y}_0 \sim p({\bm y})$, calculate $\bm y_s$.}
    \STATE {$\gamma \sim p(\gamma)$.}
    \STATE {$\epsilon \sim \mathcal{N}({\bm 0},{\bm I})$.}
    \STATE {Take a gradient descent step on \\ $\nabla_{\theta_1}\|\bm\epsilon- \bm\epsilon_{\theta_1}(\sqrt{\gamma_k}\bm y + \sqrt{1-\gamma_k}\bm \epsilon,\gamma_k,\bm y_s)\|_1$.}
    \UNTIL{converged}
  \end{algorithmic}
  \label{alg_training_a_denoising_model}
\end{algorithm}

\subsection{Conditional Translation Inference (Reverse)}

Firstly, we demonstrate that the image after translation $\bm{ \hat{y}} \sim p(\bm y_0|\bm x_0)$ could be approximated as $p(\bm y_0|\bm x_K)$.  According to the assumption that the added noise could be only estimated according to its nearest iteration and the Markov property:
\begin{align}
    p(\bm y_0|\bm x_K) =  p(\bm y_0|\bm x_K,\bm x_0)
    \label{k_01}
\end{align}
Then, according to Bayes principle:
\begin{align}
       & p(\bm y_0|\bm x_K,\bm x_0) =  p(\bm y_0|\bm x_K,\bm x_0)\frac{q (\bm x_K|\bm x_0)}{q (\bm x_K|\bm x_0)}\nonumber\\ 
    & =\frac{p(\bm y_0|\bm x_K,\bm x_0)q (\bm x_K|\bm x_0)}{q (\bm x_K|\bm y_0,\bm x_0)}=
    \frac{p(\bm y_0, \bm x_K|\bm x_0)}{q (\bm x_K|\bm y_0,\bm x_0)}\nonumber\\ 
    &= p(\bm y_0|\bm x_0) 
    \label{k_02}
\end{align}

As for the proposed inference step, diffusion models use $\bm y_K\sim \mathcal{N}(0,I)$ during the reverse process $P(\bm y_0|\bm y_K)$ (diffusion inference), and $\bm x_K$ could be considered as an approximate normal distribution. Here we assume the obtained deformation $\bm \phi$ could close the gap of shapes between source and target domain. Then because the property we have demonstrated in \ref{k_01} and \ref{k_02}. $p\left(\bm{y}_{0}|\bm{x}_{0}, \bm y_s\right)$ could be considered as $p\left(\bm{y}_{0}|\bm{x}_{K}, \bm y_s \right)$. 
Thus conditional reverse process could be described as follows:\begin{align}
    &\bm {\hat y} \sim p\left(\bm{y}_{0}|\bm{x}_{0}, \bm y_s\right)=p\left(\bm{y}_{0}|\bm{x}_{K}, \bm y_s\right)= p\left(\bm{y}_{0}|\bm{x}_{K}, \bm x_s\circ\bm \phi\right)\nonumber\\ \nonumber
     &= p\left(\bm{y}_{K-1:0} \mid \bm{x}_{K}, \bm x_s\circ \bm \phi \right)\\ 
     &=\prod_{k=1}^{K-1} p\left(\bm{y}_k \mid \bm{y}_{k-1},\bm x_s\circ \bm \phi,\gamma_k \right ) p\left(\bm{y}_{K-1} \mid \bm{x}_{K},\bm x_s\circ \bm \phi,\gamma_k \right ) 
     \label{E_all}
\end{align}
For sampling procedure, the synthetic image $\bm{\hat{y}}$ is obtained from the source image $\bm x_0$ and estimated noise  $\hat{\bm \epsilon} =\bm\epsilon_{\theta_1}(\bm {\hat y_k},\gamma_k,\bm x_s\circ\bm\phi)$,
where $\hat{\bm y}_k$ is estimated by the following iterative generation process.
\begin{equation}
\hat{\bm{y}}_{k-1}=\frac{1}{\sqrt{\alpha_k}}\left(\hat {\bm y_k}-\frac{1-\alpha_k}{\sqrt{1-\gamma_k}}\hat{\bm \epsilon}\right)+\sqrt{1-\alpha_k} {\bm z}
\end{equation}
In \ref{E_all}, similar to $p\left(\bm{y}_k \mid \bm{y}_{k-1},\bm x_s\circ \bm \phi,\gamma_k \right )$, 
 $p\left(\bm{y}_{K-1} \mid \bm{x}_{K},\bm x_s\circ \bm \phi,\gamma_k \right )$  is calculated by replacing $\bm y_K$ to $\bm x_K$, where ${\bm x}_K$ is estimated by $ q(\bm x_K|\bm x_0)$ which is shown as the following Markovian noising process.
\begin{equation}
\small
q(\bm x_K|\bm x_0) =\prod_{k=1}^{K}q(\bm x_k|\bm x_{k-1}) =\mathcal{N}(\bm x_K;\sqrt{\gamma_K}\bm x,(1-\gamma_K)I)
\end{equation}

\begin{algorithm}[h]
  \caption{Inference in $K$ iterative refinement steps}
  \begin{algorithmic}[1]
    \STATE {$\bm x_0\sim p(\bm x)$, $ \bm x_K \sim q(\bm x_K|\bm x_0).$}
    \STATE {$\hat{\bm{y}}_{K-1}=\frac{1}{\sqrt{\alpha_k}}\left(\bm{x}_K-\frac{1-\alpha_k}{\sqrt{1-\gamma_k}} \epsilon_{\theta_{1}}\left(\bm{x_s}\circ{\bm \phi}, \bm{x}_K, \gamma_k\right)\right)+\sqrt{1-\alpha_k} \bm{z}$}
    \FOR {$k = K-1, \cdots , 1$}
    \STATE {${\bm z } \sim \mathcal{N}({\bm 0},{\bm I})$ \textbf{if} $k>1$; ${\bm z =0}$ \textbf{otherwise}.}
    \STATE{$\hat{\bm{y}}_{k-1}=\frac{1}{\sqrt{\alpha_k}}\left(\hat{\bm{y}}_k-\frac{1-\alpha_k}{\sqrt{1-\gamma_k}} \epsilon_{\theta_1}\left(\bm{x_s}\circ{\bm \phi}, \hat{\bm{y}}_k, \gamma_k\right)\right)+\sqrt{1-\alpha_k} \bm{z}$}
    \ENDFOR
    \RETURN {$\hat{{\bm y}}=\hat{{\bm y}}_0$.}
  \end{algorithmic}
  \label{alg_inference_iterative_refinement_steps}
\end{algorithm}

\section{Spatial Transformation}
\subsection{Warping Operation}
Given a moving image $\bm x$ and a corresponding deformation $\bm \phi$, the resulting warped image can be denoted as $\bm x\circ \bm \phi$. For each pixel $\bm p$,  the warping operation is to re-sample the image by satisfying a subpixel location $\bm p^ \prime = \bm p + \bm u(\bm p) = \bm \phi(\bm p)$   in image $\bm x $, where $ \bm u $ is the displacement field. The deformation can be computed by adding an identity grid to $ \bm u $, i.e., $\bm \phi = \bm u + \text{\textbf{Id}}$. To facilitate gradient descent optimization,  a differentiable warping operation based on spatial transformer networks \cite{STN} 
computes the warped image $\bm x\circ \bm \phi$ as:
\begin{equation}
   \bm x \circ \bm{\phi}(\bm{p})=\sum_{\bm{q}\in \mathcal{Z}(\bm{p}^{\prime})} \bm x (\bm{q})\prod_{d \in\{m, n\}}\left(1-\left|\bm{p}_d^{\prime}-\bm{q}_d\right|\right),
\end{equation}
\noindent where $\mathcal{Z}(\bm p^ \prime)$ are the neighbours of $\bm p^\prime$ and $d$ iterates over image dimensions $\mathbb{R}^{m\times n}$.
\subsection{Scaling and Squaring}
In order to preserve the topology of anatomical structures and realize a realistic transformation, we expect the forward and backward deformations are diffeomorphic and reversible to each other. To achieve this, we leverage the static ordinary differential equation (ODE) to generate final deformations: $\partial{\bf{\bm \phi}}/\partial t  = {\bf{v}}_t({\bf{\bm \phi}}_t)={\bf{v}}_t\circ{\bf{\bm \phi}}_t$, where ${\bf{\bm \phi}}_0 = {\rm{Id}}$ represents the identity grid and ${\bf{v}}_t$ indicates the velocity field at time $t$ ($\in [0,1]$). 

To solve the ODE and obtain the series of deformations, a numerical method called scaling and squaring \cite{ss_layer} is applied to integrate stationary velocity fields (SVFs). It uses group theory to exponentiate a member of the Lie algebra $\bf v$. The properties of one-parameter subgroups indicate that $(exp((t + t^\prime) \bm v) = exp(t\bm v) \circ exp(t^\prime\bm v)) $ given any scalars $(t)$ and $(t^\prime)$, where $exp$ denoted as the exponential mapping. Thus the static composition is denoted as follows:
$\bm \phi^{1/2^{t}} = \bm \phi^{1/2^{(t+1)}}\circ \bm \phi^{1/2^{(t+1)}},$ where $\circ$ denotes the warping operation, $\bm \phi^{1/2^T}=\bm \phi_0 + \bm v/ 2^{T}$. $\bm \phi =C(\bm v)$   represents the composition process, similarly, the backward deformation can be estimated as $\bm \phi^{-1} =C(\bm v^{-1})$. We use $ T=7$ in our experiments. 


\begin{figure*}[t!]
    \centering
    \includegraphics[width=\textwidth]{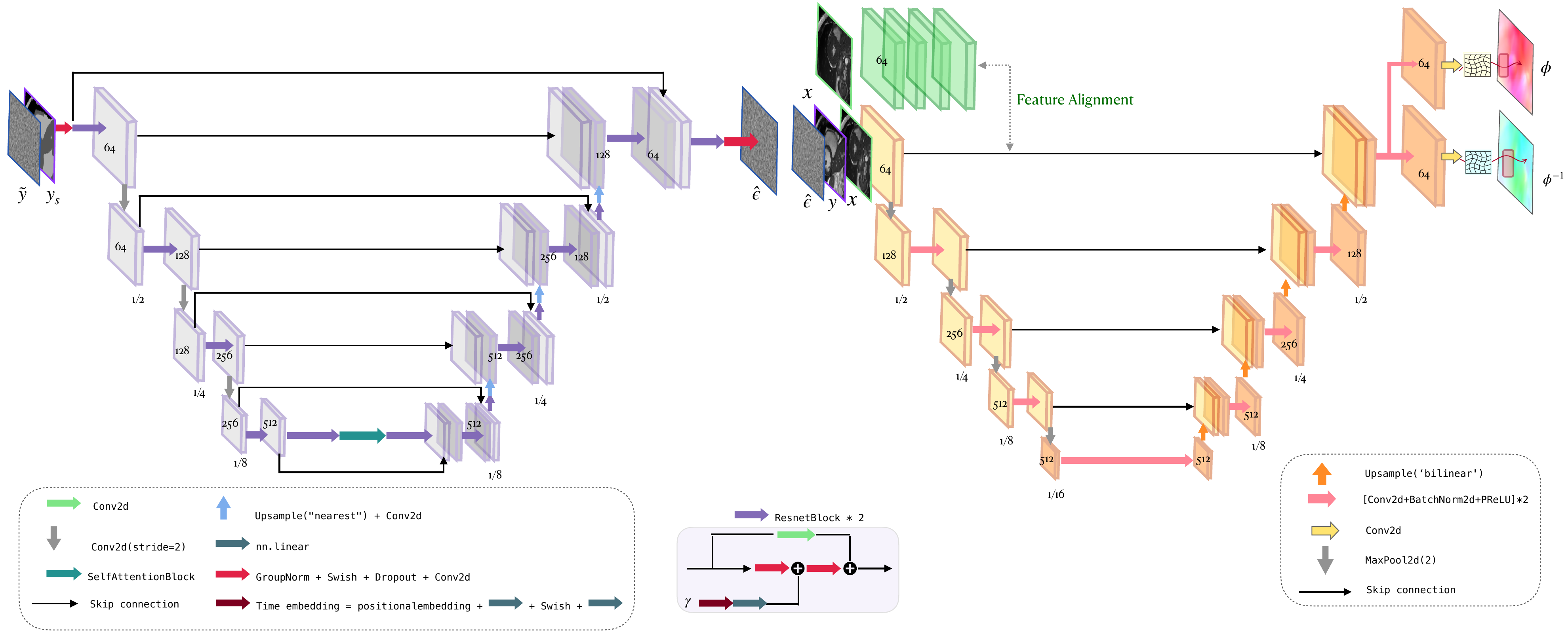}
    \caption{Illustration of noise estimation network architecture $\bm \epsilon_{\theta_1}$ (left) and U-Net architecture $\bm f_{\theta_2}$ (right). The number of convolution channels is labelled on the feature maps. The number placed below the feature maps represents the ratio of image dimensions that are compared with the original input size. In network $\bm \epsilon_{\theta_1}$, the noised target image $\tilde{\bm y}$ and are gathered with the conditional structure map $\bm y_s$ as the input pair and fed into network $\bm \epsilon_{\theta_1}$ to learn the encoded image feature $\hat{\bm \epsilon}$. The network $\bm f_{\theta_2}$ takes three inputs which include the source image $\bm x$, the target image $\bm y$ and the encoded image feature $\hat{\bm \epsilon}$, then estimates the forward and inverse velocity fields $\bm \phi$ and $\bm \phi^{-1}$. }
    \label{fig:N_A}
\end{figure*}

\subsection{Mutual Information and Smoothness Term}

As we mentioned in the manuscript, the final objective of the spatial transformation module includes an energy term and a regularization term. The energy term, i.e., data similarity, is desired to maximize the mutual information between a warped moving image and a fixed image. However, the computation of mutual information is non-differentiable as it requires to construct a histogram, to achieve a differentiable mutual information loss, we adopt the implementation from \cite{qiu2021learning}, where they used a differentiable Parzen window to map the image intensities to histogram. Particularly, the mutual information loss $L_{MI}(\bm y, \bm x \circ\bm \phi)$ is described as follows.
\begin{equation}
    L_{MI}(\bm y, \bm x \circ\bm \phi)=\frac{H(\bm y)+H(\bm x \circ\bm \phi)}{H(\bm y, \bm x \circ\bm \phi)},
\end{equation}
where $H(\bm y)$ and $\bm x \circ\bm \phi$ denote the marginal entropies,
and $H(\bm y, \bm x \circ\bm \phi)$ represents the joint entropy. Specifically, the entropy is defined by the intensity distribution $p(I_x)$ by $H( I_x) = -\int_{x}p(I_x)\ln{(p(I_x))}d I_x$. To estimate the intensity distribution, a differentiable Parzen window \cite{thevenaz2000optimization} is used to calculate the intensity distribution $p(I_x, I_y)$, where $I_x$ and $I_y$  denote intensity values. $p(I_y)=\sum_{I_x \in L_{\bm x\circ \bm \phi}}p(I_x, I_y)$ and $p(I_x)=\sum_{I_y \in L_{\bm y }}p(I_x, I_y)$.

The smoothness regularization is based on first-order gradient implemented using the finite differences, similar to \cite{dalca2018unsupervised,jia2022u}.

\section{Network Architecture} 

\noindent{\bf{Noise estimation network architecture.}}
As shown in the left diagram of \ref{fig:N_A}, the encoder blocks and decoder blocks in $\bm \epsilon_{\theta_1}$ both have four layers. The residual blocks are employed in each layer to incorporate the information of the embedded noise level $\gamma$. A basic block (red arrow) in residual blocks consists of group normalization \cite{Wu_2018_ECCV}, the swish function\cite{ELFWING20183}, the dropout layer\cite{JMLR:v15:srivastava14a} and convolution layers.   The preceding layer of the encoder halves the dimension of the image using the convolution operation with 2 strides. Similarly, the encoder blocks up-sample image dimensions using the nearest interpolation with scale factor 2. The outputs of each encoder layer are fed into the corresponding decoder layer to enhance fine-grained details for contributing the final reconstructed outputs via the skip connection. Additionally, self-attention block (lake blue arrow) \cite{attention} is capable of capturing the context of spatial positions. 
\medskip\\
\noindent{\bf{U-Net architecture.} } Similarly, the architecture of $\bm f_{\theta_2}$ is a U-shape-like network which is comprised of four encoder blocks and four decoder blocks. As shown in right diagram of \ref{fig:N_A}, the architecture is adapted from VoxelMorph \cite{dalca2018unsupervised} with 64 start channels for 2D inputs. We modified the final convolution layer to two output streams with the same convolution kernel [64, 2, 1] for leaning both forward and inverse velocity fields. Additionally, we customised the input switch layer by adding a 4-layer auxiliary convolution block with the same kernel size [64, 64, 3] to align the input features of $\bm y$ and $\hat{ \bm  \epsilon}$.
\medskip

\begin{figure}[h!]
    \centering
  \includegraphics[width=0.8\linewidth]{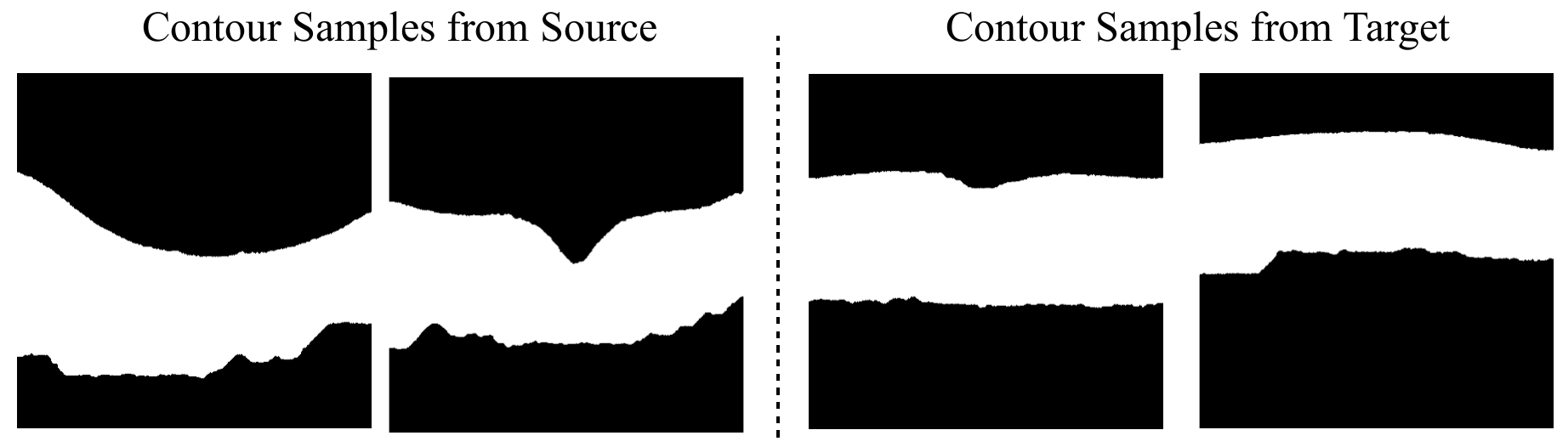}
    \caption{Contour samples from source and target datasets.}
    \label{fig:samples_contour}
\end{figure}

\noindent{\bf{Contour samples} }  Since only foreground and background regions are involved in guiding the contour shapes, the clustering task remains relatively simple. To clarify this, we provide visualizations of the clustering results ($x_s$ and $y_s$) in Figure~\ref{fig:samples_contour}, which show that the clusters capture basic structural separation sufficient to guide noise estimation and spatial transformation. \\

\begin{figure}[h!]
    \centering
    \includegraphics[width=\linewidth]{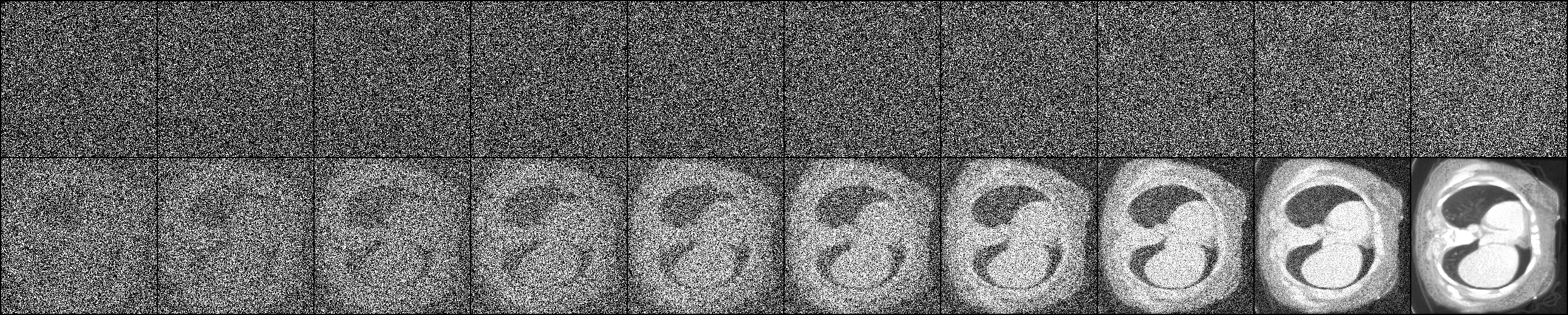}
    \caption{Intermediate denoising steps sampled from noised source image.}
    \label{fig:long_mri}
\end{figure}

\noindent{\bf{Intermediate denoising steps} }  Figure~\ref{fig:long_mri}, which visualizes the denoising trajectory and demonstrates how the image is progressively reconstructed.\\

\noindent{\bf{Baseline Methods} } 

\noindent{{\bf CycleGAN}\cite{zhu2017unpaired} applies a cycle consistency loss into two adversarial losses to learn two mapping functions $X \to Y$ and $Y \to X$ which constrains satisfied $F(G(X)) \approx X$. Two generators $\{G,F\}$ and discriminators $\{D_X,D_Y\}$ are required to form a cycle consistency loss denoted as \(\mathbb{E}_{x\sim p(x)} {\lVert {F (G(x)) - x} \rVert}_1 +\mathbb{E}_{y\sim p(y)} {\lVert {G(F(y)) - y} \rVert}_1\). }\medskip\\
\noindent{{\bf MUNIT}\cite{huang2018multimodal} use disentangled representations to encode the source and target images into two latent spaces that contain domain-invariant and domain-specific attributes, then trains image reconstruction losses and latent reconstruction losses for content features and style features.}\medskip\\
\noindent{{\bf DualGAN}\cite{yi2017dualgan} uses a primal-dual manner to allow translating two sets of unlabeled images between two different domains by minimising two reconstruction losses $\lVert {G_A(G_B(v,z^\prime),z)-v} \rVert$ and $\lVert {G_B(G_A(u,z),z^\prime)-u} \rVert$.
}\medskip\\
\noindent{{\bf NAGAN} \cite{zhang2019noise}} uses two discriminators (style and content) to translate images from source to target domain. The translated styles are obtained from the shallow layers of the style discriminator by aligning Gram matrix of the translated image and the target one.\medskip\\
\noindent{{\bf StarGAN v2} \cite{choi2020stargan} incorporates an auxiliary classifier to achieve multi-domain translation translation using a single discriminator by considering both sources and domain labels, $D:x \to \{D_{src}(x), D_{cls}(x)\}$. The trained generator $G$ translates a source image $x$ into a conditional image $y$ associated with target domain class labels $c$, which denotes as $G(x, c) \to y$.}\medskip\\
\noindent{{\bf DDIB} \cite{su2022dual} encode images on the source and target domain using two diffusion models which connecting the noising images distributions through Schrödinger bridge. Then it synthesises a target image with a corresponding encoded source's noise image by solving reverse ODE on the trained target model.}\medskip\\
\noindent{{\bf UNSB} \cite{kim2024unpaired} reformulates unpaired I2I translation as a Schrödinger bridge problem solved via neural stochastic differential equations, unifying diffusion modeling and optimal transport. It learns forward and backward dynamics to map between source and target distributions.}\medskip\\
\noindent{{\bf Arar et al.} \cite{Arar_2020_CVPR} accomplished multi-modal image registration by applying registration to translated images. This method aims to minimize dissimilarities between moving and fixed images across different modalities by a straight forward combination of CycleGAN and registration network.

}
\subsection{User Study Details}
For our clinical user study, we expected three clinicians to view and mark the experimental results.  We asked the clinician three questions on  disease progression aspect \textit{i.e.} the disease (\textit{e.g.} retinal AMD Pathology Drusen), the fidelity of the generated images, and satisfaction with the achieved traceability. More specifically, the questions are listed as follows: 1) Are generated lesions on translated images reasonable (locates on a proper position and have realistic appearances and sizes)? 2) Are the generated images realistic?  3) Do outputs have correspondences of source images? Finally, we set the scores from 1 to 5 to grade the translated images, where a lower score denotes better result.

\bibliographystyle{IEEEtran}
\bibliography{main.bib}

\end{document}